\documentclass{article}

\usepackage[nonatbib,final]{neurips_2022}

\usepackage[utf8]{inputenc} 
\usepackage[T1]{fontenc}    
\usepackage{hyperref}       
\usepackage{url}            
\usepackage{booktabs}       
\usepackage{amsfonts}       
\usepackage{nicefrac}       
\usepackage{microtype}      
\usepackage{xcolor}         

\usepackage{microtype}
\usepackage{graphicx}
\usepackage{caption}
\usepackage{subcaption}
\usepackage{amsmath,amssymb}
\usepackage[normalem]{ulem}
\usepackage{xstring}
\usepackage{mathtools, nccmath}
\usepackage{xpatch}
\usepackage{amsthm}
\usepackage{siunitx}
\usepackage{hyperref}
\usepackage{wrapfig}

\usepackage{algorithm} 
\usepackage{algpseudocode} 

\newcommand{\method}[1]{\texttt{ClonEx-SAC}}

\title{Disentangling Transfer in Continual Reinforcement Learning}

\author{%
  Maciej Wołczyk\thanks{equal contribution} \\
  Jagiellonian University\\
  Kraków, Poland \\
  \texttt{maciej.wolczyk@doctoral.uj.edu.pl} \\
  \And
  Michał Zając$^*$ \\ 
  Jagiellonian University\\
  Kraków, Poland \\
  \texttt{emzajac@gmail.com} \\
  \AND
  Razvan Pascanu \\
  DeepMind \\
  London, UK \\
  \texttt{razp@google.com} \\
  \And
  Łukasz Kuciński \\
  Polish Academy of Sciences \\
  Warsaw, Poland \\
  \texttt{lkucinski@impan.pl} \\
  \And
  Piotr Miłoś \\
  Ideas NCBR,\\
  Polish Academy of Sciences, \\
  deepsense.ai\\
  Warsaw, Poland \\
  \texttt{pmilos@impan.pl} \\
}

\begin{document}

\maketitle

\begin{abstract}
The ability of continual learning systems to transfer knowledge from previously seen tasks in order to maximize performance on new tasks is a significant challenge for the field, limiting the applicability of continual learning solutions to realistic scenarios.
Consequently, this study aims to broaden our understanding of transfer and its driving forces in the specific case of continual reinforcement learning. We adopt SAC as the underlying RL algorithm and Continual World as a suite of continuous control tasks. We systematically study how different components of SAC (the actor and the critic, exploration, and data) affect transfer efficacy, and we provide recommendations regarding various modeling options. The best set of choices, dubbed \method{}, is evaluated on the recent Continual World benchmark. \method{} achieves $87\%$ final success rate compared to $80\%$ of PackNet, the best method in the benchmark. Moreover, the transfer grows from $0.18$ to $0.54$ according to the metric provided by Continual World.

\end{abstract}

\section{Introduction}\label{sec:introduction}

The ability of continual learning (CL) systems (\cite{HADSELL20201028, khetarpal2020continual}) to {utilize} knowledge from previously seen tasks in order to maximize {transfer on the current task} is a significant challenge for the field.  Achieving progress in this area would bring benefits both for real-life applications and multiple machine learning domains \cite{DBLP:conf/nips/KrizhevskySH12,he2016deep, DBLP:conf/nips/VaswaniSPUJGKP17,bert}, including reinforcement learning (RL), as advocated in \cite{wolczyk2021continual}. In particular, it would constitute a critical step towards making efficient lifelong learning agents a reality.

The goal of this paper is to expand our understanding of transfer and its driving factors in continual reinforcement learning (CRL). As the underlying RL algorithm, we assume soft actor-critic (SAC), see \cite{DBLP:journals/corr/abs-1812-05905}, and use Continual World \cite{wolczyk2021continual} as the suite of continuous control environments. We systematically study the critic and actor networks, the key components of SAC, with regard to their influence on transfer. Similarly, we measure the impact of various choices regarding exploration and buffer data usage.
The low-level mechanisms of transfer are not yet fully understood even in the supervised learning case \cite{neyshabur2020being}. To the best of our knowledge, our work is the first one that undertakes a comprehensive study of this important topic in RL. To this end, we proceed in two stages: exploring a two-task setting and a full continual learning scenario.

  \begin{wrapfigure}{R!}{0.6\linewidth}
    \vskip -0.2in
    \begin{center}
        \centerline{
        \includegraphics[width=.58\columnwidth]{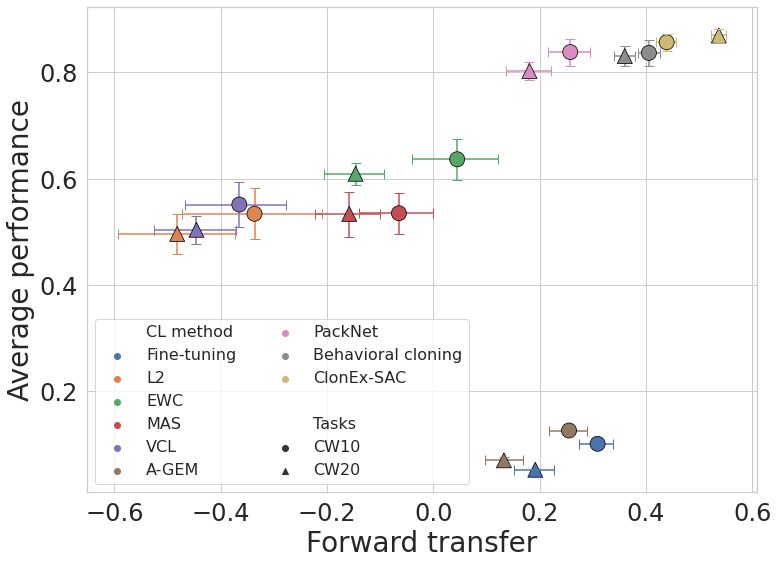}
        }
        \caption{\small Performance of the \method{} method compared with competitive baselines, on CW10 and CW20 task sequences. Average performance and forward transfer are shown, together with 90\% bootstrap confidence intervals.}
        \label{fig:main_clonex_results}
    \end{center}
    \vskip -0.3in
  \end{wrapfigure}

We start by investigating a simplified two-tasks setting in Section \ref{sec:transfer_experiments}. This allows us to leave out the impact of forgetting, as well as limit the choices regarding exploration and data handling. We use $100$ pairs of robotic tasks from the Continual World benchmark.
Here, we render two key observations: 1) the role of the critic is the most important for transfer, while exploration and actor play smaller, but non-negligible, parts; 2) contributions of the individual components are mostly independent. Additionally, we show that the concept of feature reuse which is often utilized to explain supervised transfer learning \cite{neyshabur2020being,raghu2019rapid} might not be directly applicable in RL.

In Section \ref{sec:transfer_in_sequences}, we aim to understand new effects which emerge for the full continual learning scenario, typically in longer sequences. In CL context, we need to take into account forgetting, being mindful of the fact that existing CL methods often favor mitigating forgetting at the expense of transfer, see \cite{wolczyk2021continual}. Main results include 1) reusing policies from previous tasks for exploration considerably improves performance; 2) behavioral cloning  to rehearse past tasks is beneficial for both average performance and forward transfer, outperforming other considered methods; 3) regularizing the critic typically does not help for the performance of CL methods.

The result of our comprehensive analysis is a set of general recommendations. We also determine the combination of design choices that outperforms all other options, dubbed \method{}. This method utilizes behavioral \textit{cloning} to mitigate catastrophic forgetting for the actor. Moreover, at the beginning of each task, \method{} queries all previous policies, the best of which generates initial \textit{exploration} data. \method{} achieves $87\%$ final success rate compared to $80\%$ of PackNet, the best method in the Continual World benchmark, see Figure~\ref{fig:main_clonex_results}. Importantly, we observe a sharp transfer increase from $0.18$ to $0.54$ in the metric provided in the benchmark. Notably, the value of forward transfer closely matches the reference forward transfer adjusted for exploration, which is a soft upper bound for transfer, as introduced in \cite{wolczyk2021continual}.

\section{Related work}
Continual learning algorithms are often categorized into three classes: regularization-based e.g. \cite{DBLP:conf/eccv/AljundiBERT18, DBLP:journals/corr/KirkpatrickPRVD16,DBLP:conf/iclr/NguyenLBT18}, parameter isolation e.g. \cite{DBLP:conf/cvpr/MallyaL18} and rehearsal methods e.g. \cite{DBLP:conf/iclr/ChaudhryRRE19,DBLP:journals/corr/abs-1902-10486};
see also CL survey papers \cite{de2019continual, HADSELL20201028,DBLP:journals/nn/ParisiKPKW19}. 
\cite{DBLP:journals/corr/abs-2107-12808,DBLP:journals/corr/abs-1909-07528} advocate the need to develop CL methods suitable for reinforcement learning training 
as a necessary step towards learning artificial intelligent agents to operate in open-ended and changing environments.
\cite{khetarpal2020continual} provides a detailed review of this combination and a taxonomy of possible setups. \cite{wolczyk2021continual} proposes a sequence of robotic tasks as a benchmark, comparing popular CL methods adapted to RL and advocating for putting more emphasis on transfer.

The authors of \cite{kaplanis2018continual} show how the synaptic Benna-Fusi model can be added on top of value-based RL methods to mitigate forgetting at both intra- and inter-task scales. A simple approach to cloning policies from previous tasks is employed in \cite{DBLP:journals/corr/abs-1907-05855}, and a similar replay strategy has been used in \cite{rolnick2019experience}.  \cite{DBLP:journals/corr/abs-2106-02940} tackles the case when task boundaries are not provided. Although most of the research is concerned with model-free continual reinforcement learning, an approach to model-based continual RL was presented in \cite{huang2021continual}.

Transfer learning, which focuses on the reuse of machine learning models, has been extremely successful recently. 
In computer vision, convolutional neural networks \cite{DBLP:conf/nips/KrizhevskySH12,he2016deep} and vision transformers \cite{dosovitskiy2020image} pre-trained on large datasets can be repurposed and fine-tuned on the target task. Modern transformer-based models \cite{DBLP:conf/nips/VaswaniSPUJGKP17,bert} trained on large natural language corpora turned out to be very flexible and can be adapted to diverse downstream tasks with surprising efficiency \cite{DBLP:journals/corr/abs-2201-12122, lu2021pretrained}. General surveys of transfer learning techniques are provided in \cite{zhuang2019comprehensive,10.1007/978-3-030-01424-7_27}. 
Interestingly, recent research \cite{NEURIPS2020_0607f4c7,DBLP:journals/corr/abs-1910-04867}  suggests that there are still some gaps in our understanding of transfer learning.
\cite{NEURIPS2020_0607f4c7} analyzes the low-level reasons for transfer, exhibiting surprising phenomena such as transfer between datasets with permuted images. \cite{DBLP:journals/corr/abs-1910-04867} performs large-scale experiments investigating representation transfer in a wide variety of visual tasks. 

In reinforcement learning scenarios, the structure of the underlying MDP can be exploited to facilitate the transfer. \cite{taylor2009transfer,zhu2020transfer, taylor2007transfer,talvitie2007experts} present methods on how to find and use mappings between different domains.  \cite{ng1999policy,brys2015policy} apply reward function reshaping. \cite{mehta2008transfer,steccanella2020hierarchical} achieve transfer by means of high-level skills and hierarchical RL. Other lines of work exploit the model structure \cite{rusu2016progressive,fernando2017pathnet} or enforce modularity \cite{andreas2017modular,devin2017learning}. In this work, we aim to complement these studies, by focusing on the benefits of reusing neural network parameters, and other choices that exploit the RL structure, like exploration and data rehearsal.

\section{Background} 
\label{ref:background}

\subsection{Continual learning and reinforcement learning}
Continual learning tackles the problem of learning in non-stationary settings \cite{de2019continual}. Typically, the solution is expected to perform well on all encountered tasks, although various metrics expressing different requirements are formulated.
The popular CL desiderata include reducing the \emph{forgetting} on previous tasks and increasing the \emph{forward transfer} on the new tasks, i.e.\ speeding up the learning by reusing knowledge from previous tasks \cite{DBLP:journals/corr/abs-1805-09733, wolczyk2021continual}. Other desiderata focus on limiting resources, such as the number of samples, computation time, model size, or additional memory size. These requirements are often conflicting, so usually some trade-offs have to be made \cite{HADSELL20201028, wolczyk2021continual,DBLP:journals/nn/ParisiKPKW19}.

Combining CL with RL adds another layer of complexity. In this work, we focus on the SAC algorithm \cite{DBLP:journals/corr/abs-1812-05905}, which is often considered to be the method of choice for continuous control RL \cite{metaworld,yu2020gradient,shao2019survey}. 
As an actor-critic algorithm, it is based on the interplay between its two parts, see Section~\ref{sec:sac}. This is a fairly complicated algorithmic setup, which presents a number of challenges when used jointly with CL.

In particular, since the optimization of the critic and actor networks is intertwined, it is hard to understand and decouple the impact of individual components. Additionally, because of this interplay, training biases get easily exacerbated, often leading to inferior performance or even a collapse. Another complication is that the training objectives for the actor and the critic are different. The critic
    minimizes the Bellman error which is known to be a fragile objective \cite{fujimoto2022i} susceptible to training biases and might correlate poorly with the value error (which we would like to minimize). As the actor optimizes over predictions of the critic, it might also suffer from these problems, even if less directly. 
  Finally, since the policy and the data we see change during the training, there is an inherent distribution shift present, even within a single task.

\subsection{SAC}
\label{sec:sac}

In our study, we focus on soft actor-critic (SAC)~\cite{haarnoja2018soft}, an off-policy actor-critic RL algorithm, based on the maximum entropy principle. 
\textbf{The critic} strives to approximate the entropy-corrected $Q$-function under the current policy, optimizing the Bellman error. \textbf{The actor} tries to find actions that maximize the $Q$-function. \textbf{The replay buffer} holds the seen experience and provides data for the actor and critic updates at each learning step. \textbf{The exploration policy} is used to gather data at the beginning of each task for a set number of $K$ steps. By default, in most SAC implementations, this means sampling actions uniformly over the action space.

\subsection{Continual World}\label{sec:ContiualWorld}

We perform our experiments on the Continual World \cite{wolczyk2021continual} benchmark. It contains a set of realistic robotic tasks, where a simulated Sawyer robot manipulates everyday objects. The structure of the observation and action spaces remains the same between the tasks; an observation is a $12$-dimensional vector describing the coordinates of the robot's gripper and relevant objects. The $4$-dimensional action space describes the gripper movement. In training, a dense reward function is used to make the tasks solvable; in evaluation, the binary success metric is used to indicate whether the desired goal has been reached.
The tasks are arranged in sequences and training in each task lasts for 1M steps. CW10 sequence contains $10$ different tasks arranged in a fixed order. CW20 consists of CW10 repeated twice, allowing to measure how much knowledge can be transferred in case of task repetitions. We use both CW10 and CW20 in our evaluations, as well as shorter sequences containing pairs of tasks from CW10.

\subsection{Metrics}

Following standard practice in continual learning literature, we report average performance and forgetting metrics. We also measure transfer as defined in \cite{wolczyk2021continual}. Below we briefly recall these three metrics. Assume $p_i(t)\in[0,1]$ to be the performance (success rate) of task $i$ at time $t$, and that each of the $N$ tasks is trained for $\Delta$ steps, so the total number of steps is $T=N\cdot \Delta$.

\paragraph{Average performance} The average performance at time $t$ is defined as 
    $\text{P}(t) := \frac{1}{N} \sum_{i=1}^{N} p_i(t)$.\label{eq:performance}
Its final value, $P(T)$, is a scalar summary of the performance and is presented in the result tables.

\paragraph{Forward transfer} The forward transfer is computed as a normalized area between the training curve of the measured run and the training curve of a reference curve from training from scratch. Let us denote by $p_i^b\in[0,1]$ the reference performance. Then the forward transfer on task $i$, $FT_i$, is defined as
\begin{align*}
    \text{FT}_i    & := \frac{\text{AUC}_i-\text{AUC}_i^b}{1-\text{AUC}_i^b},\quad                        
    \text{AUC}_i   := \frac{1}{\Delta}\int_{(i-1)\cdot\Delta}^{i \cdot \Delta} p_i(t) \text{d}t, \quad 
    \text{AUC}_i^b  :=  \frac{1}{\Delta}\int_{0}^{\Delta} p_i^b(t) \text{d}t.
\end{align*}
The average forward transfer for all tasks, $\text{FT}$, is defined as
    $\text{FT} = \frac{1}{N}\sum_{i=1}^N \text{FT}_i$.

\paragraph{Forgetting} For the task $i$, one can measure a drop in performance after the end of learning on this task as
    $F_i = p_i(i \cdot \Delta) - p_i(T).$ \label{eq:forgetting}
Forgetting metric is defined as $F = \frac{1}{N}\sum_{i=1}^N F_i$.

\subsection{Experimental setup}
\label{sec:experimental_setup}
We follow the experimental setup from \cite{wolczyk2021continual}. The actor and the critic are implemented as two separate MLP networks, each with 4 hidden layers of 256 neurons. We refer to the 4 hidden layers as the \textit{backbone} and the last output layer as the \textit{head}. By default, we assume the \textit{multi-head} (MH) setting, where each task has its separate output head, but we also consider the \textit{single-head} (SH) setting, where only a single head is used for all tasks. The SAC exploration phase takes $K =10\text{k}$ steps. All experiments in this paper were performed with $10$ different seeds unless noted otherwise. We compute $90\%$ confidence intervals through bootstrapping.
More details on the experimental setup can be found in Appendix~\ref{sec:setting_appendix}.

\section{Transfer in isolation } \label{sec:transfer_experiments}

In this section, we  study \textit{what enables transfer between RL tasks}. We assume a two-task setting, where we measure the forward transfer from the first to the second task, disregarding issues specific to continual learning (e.g. forgetting), which we defer to the next section. We utilize all $100$ pairs of CW10 tasks, see Section \ref{sec:ContiualWorld}, to evaluate the impact of \textit{critic, actor, and exploration} given by SAC. 

We will say that the actor or the critic are \textit{carried over} (from the previous tasks) if their parameters are reused as the initialization in the next task; otherwise, the parameters are re-initialized. We also refer to the exploration policy as being carried over, if we use the policy from the previous task (or tasks) to gather the data during the first $10$k steps of the SAC exploration phase (see Section \ref{sec:sac}); otherwise, a uniform random exploration policy is being used. We use both multi-head (MH) and single-head (SH) settings, with the former being default.

\begin{figure}[t]
    \begin{subfigure}[b]{0.325\textwidth}
        \centering
        \includegraphics[width=\textwidth]{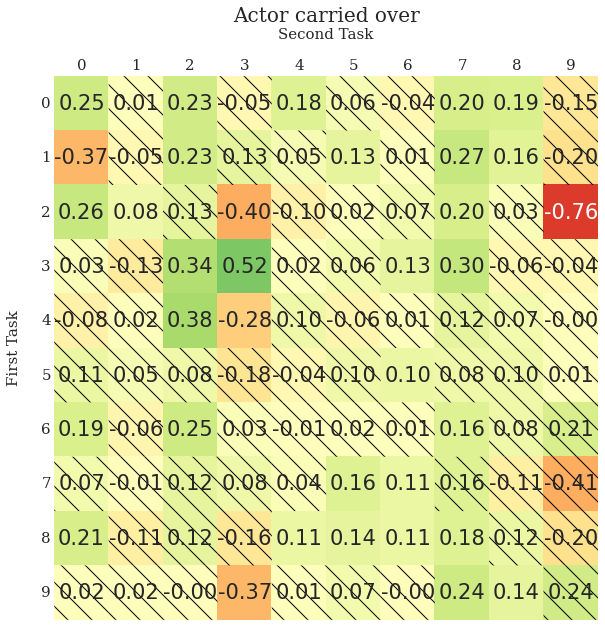}
        \caption{Actor carried over}
        \label{fig:transfer_actor}
    \end{subfigure}
    \hfill
    \begin{subfigure}[b]{0.325\textwidth}
        \centering
        \includegraphics[width=\textwidth]{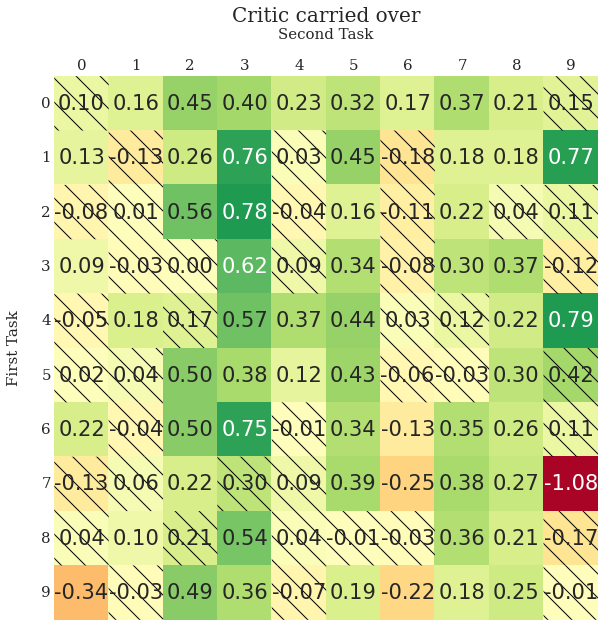}
        \caption{Critic carried over}
        \label{fig:transfer_critic}
    \end{subfigure}
    \hfill
    \begin{subfigure}[b]{0.325\textwidth}
        \centering
        \includegraphics[width=\textwidth]{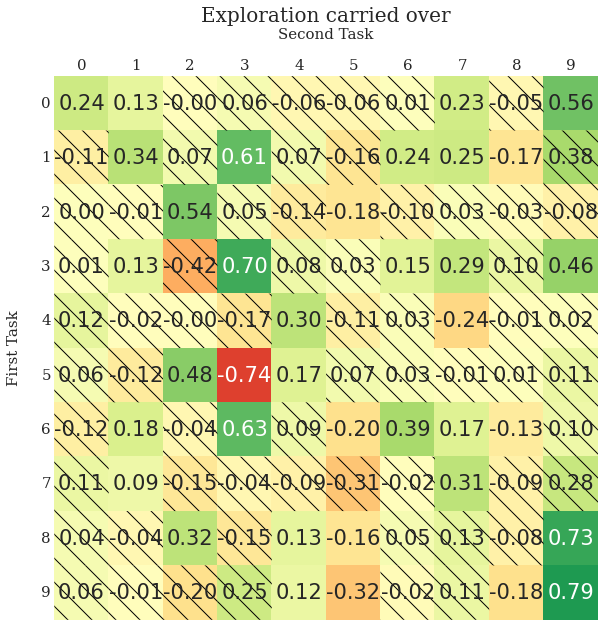}
        \caption{Exploration carried over}
        \label{fig:transfer_exploration}
    \end{subfigure}
\caption{\small The effect of carrying over different components on the performance on pairs of tasks from CW10. We shade an entry if the $90\%$ confidence interval contains $0$, indicating that we cannot be sure whether the component which was carried over makes a difference. }
\label{fig:example_transfer_matrices}
\end{figure}

Figure \ref{fig:example_transfer_matrices} illustrates the impact of the individual components on transfer for each pair. The $(i,j)$-th entry in the matrix contains the forward transfer value when carrying over components from task $i$ to task $j$.
Table~\ref{tab:transfer_basics} presents the aggregated statistics from the matrices given in Figure \ref{fig:example_transfer_matrices}:  the average FT  (including and excluding the diagonal), and the number of pairs with positive, negative, and neutral FT\footnote{We say that a pair has positive (resp. negative) FT if  the corresponding confidence interval is above (resp. below) $0$. Otherwise, we mark it as neutral.}.
Table \ref{tab:transfer_multiple} 
reports the transfer properties for all possible combinations of components present in Table \ref{tab:transfer_basics},  
omitting single-head critic (since it performs worse in Table \ref{tab:transfer_basics}).

\subsection{Carrying over SAC components } \label{sec:basic_transfer}

From the results presented above, we draw two key observations. First, the role of the critic is the most important for FT, while exploration and actor play smaller, but non-negligible, parts. Second, the components are "transfer independent", in the sense that the transfer of the combination of the components is close to the sum of transfers yielded by each component alone.

The evidence for the first finding is presented in detail for each pair in Figure \ref{fig:example_transfer_matrices} and summarized in Table \ref{tab:transfer_basics}. 
More precisely, the average forward transfer across all pairs attributed to carrying over of the critic equals $0.2$ (resp. $0.15$) for MH (resp. SH) setup. This separates the critic from the actor and exploration, which yield (for the default MH setup) $0.06$ and $0.09$, respectively.

The importance of the critic is further emphasized by showing that restraining its learning capabilities, even when the weights are initialized to the parameters learned in the previous task, negatively impacts FT. This is shown in the last row of Table \ref{tab:transfer_basics}, where only the critic’s head is allowed to train, while the body of the network is kept frozen and carried over from the previous task. This result goes against our understanding of transfer in supervised learning, where feature reuse is a common technique  (e.g. in vision \cite{neyshabur2020being,raghu2019rapid}). 
However, the deterioration in FT can be explained by RL-specific factors. Namely, freezing the backbone can hinder both the policy training (since the mechanics of SAC intertwines actor and critic) and the critic training (due to inflated Bellman errors).

As to the second finding, i.e. the "transfer independence" of the components, the results of the underlying analysis are presented in Table \ref{tab:transfer_multiple}. 
We observe that the reported FT for the combination of components follows closely the sum of FTs for individual components (reported in Table \ref{tab:transfer_basics}). Furthermore, we observe that including all the components results in the highest transfer of $0.35$.

There is a couple of remaining interesting observations. First, Figure \ref{fig:example_transfer_matrices} exhibits several vertical patterns, meaning that transfer depends more on the second task. Second, the effect on transfer increases on the diagonal, when the exploration is carried over. This seems reasonable since the policy in the new task is initialized to the already learned policy on the same task. 
Finally, resetting the head (MH setup) is beneficial in the case of the critic, while it hurts the actor.

\begin{table*}
\small
\centering

\caption{\small Summary of the transfer statistics from the transfer matrices when transferring only a single component. FT and FT (no diag) represent average forward transfer across all pairs with and without considering the diagonal (transfer from a task to the same task), respectively. Subsequent columns denote the number of pairs with the positive, negative, and neutral transfer. }

\begin{tabular}{lccrrr}
\toprule
                        name &                        FT &              FT (no diag) &  \# pos. &  \# neg. &  \# neutral \\
\midrule
Actor (MH) &  0.06 \tiny{[0.03, 0.10]} &  0.05 \tiny{[0.01, 0.09]} &       30 &        5 &          65 \\
Critic (MH) &  0.20 \tiny{[0.17, 0.23]} &  0.19 \tiny{[0.16, 0.23]} &       54 &        5 &          41 \\
Exploration &  0.09 \tiny{[0.06, 0.13]} &  0.06 \tiny{[0.03, 0.10]} &       28 &        9 &          63 \\
\midrule
Actor (SH) &  0.12 \tiny{[0.09, 0.15]} &  0.12 \tiny{[0.09, 0.15]} &       37 &        1 &          62 \\
Critic (SH) &  0.15 \tiny{[0.12, 0.18]} &  0.13 \tiny{[0.10, 0.16]} &       41 &       19 &          40 \\
\midrule
Critic (train only head) & -1.29 \tiny{[-1.33, -1.25]} & -1.30 \tiny{[-1.35, -1.26]} &        0 &      100 &           0 \\
\bottomrule
\end{tabular}

\label{tab:transfer_basics}
\end{table*}

\begin{table*}[t]
\small
\centering

\caption{\small Summary of transfer statistics when multiple components are carried over. We observe that impact of each component is largely independent of other components. That is, FT when carrying over multiple components is close to the sum of FT when carrying over each of them separately. }

\begin{tabular}{lccrrr}
\toprule
name &                       FT &             FT (no diag) &  \# pos. &  \# neg. &  \# neutral \\
\midrule
Actor (MH) + Critic (MH) & 0.27 \tiny{[0.24, 0.30]} & 0.25 \tiny{[0.22, 0.29]} &       58 &        4 &          38 \\
Actor (SH) + Critic (MH) & 0.29 \tiny{[0.26, 0.32]} & 0.28 \tiny{[0.25, 0.31]} &       59 &        2 &          39 \\
Actor (MH) + Exp. & 0.16 \tiny{[0.12, 0.20]} & 0.14 \tiny{[0.10, 0.18]} &       39 &        3 &          58 \\
Actor (SH) + Exp. & 0.21 \tiny{[0.17, 0.24]} & 0.18 \tiny{[0.15, 0.22]} &       53 &        0 &          47 \\
Critic (MH) + Exp. & 0.30 \tiny{[0.27, 0.33]} & 0.28 \tiny{[0.25, 0.31]} &       64 &        2 &          34 \\
\midrule
Actor (SH) + Critic (MH) + Exp. & 0.36 \tiny{[0.33, 0.38} & 0.33 \tiny{[0.29, 0.36]} &       68 &        0 &          32 \\
Actor (MH) + Critic (MH) + Exp. & 0.35 \tiny{[0.31, 0.38]} &    0.32 \tiny{[0.29, 0.36]} &       70 &        1 &          29 \\
\bottomrule
\end{tabular}

\label{tab:transfer_multiple}
\end{table*}

\section{Transfer in continual learning}
\label{sec:transfer_in_sequences}

In Section \ref{sec:basic_transfer}, we focused on direct transfer in the two-task setting. Now, we move to the full continual learning scenario, which brings two main differences: 1) we measure the performance of all tasks in the sequence, so forgetting now plays a significant role; 2) typically, we consider longer sequences of tasks of length $10$ and $20$ (CW10 and CW20, respectively). For longer sequences, forgetting and transfer may have complex mutual interactions. To reduce forgetting, CL methods usually apply some kind of regularization to the model, which in turn may be harmful to transfer. On the other hand, transfer benefits from accumulated knowledge -- if forgetting is not mitigated, there might be nothing to transfer from.

We will investigate three main themes. The first one is reusing previous policies for exploration. For long sequences, there are multiple design choices available compared to the two-task scenarios. Secondly, we investigate CL with data reuse, an approach successful in supervised learning.
We show that the CRL setup is more complex and requires careful investigation. Finally, given the importance of the critic for transfer (see Section \ref{sec:transfer_experiments}), we study whether the critic should be regularized or not, and conclude that typically, the answer is negative.

We study these issues in conjunction with various CL methods: Fine-tuning, Perfect memory, EWC, PackNet, L2, A-GEM, MAS, and VCL. These are standard CL approaches adopted and tested in the RL setting \cite{wolczyk2021continual}, see details in Appendix~\ref{sec:methods_appendix}. We note that CL methods used here are mostly successful in mitigating forgetting; in this section, we report average performance and forward transfer, deferring forgetting to the Appendices \ref{sec:results_appendix} and \ref{sec:experiments_general_cl_appendix}.

\subsection{Exploration} \label{sec:exploration}
When using the SAC algorithm, at the beginning of each task, there is a short period of exploration with a random policy, see Section~\ref{sec:sac}. The experiments in Section~\ref{sec:basic_transfer} showed that the transfer increases if the policy from the previous task is used instead. 
Now, we pass from two-task scenarios to longer ones, and analyze the following options for choosing exploration policy, which we call: \textit{random, preceding, uniform-previous}, and \textit{best-return}, and define them as follows.  In the first task, we always use a random policy, and  
assume that the tasks are numbered from $1$ to $N$.

Consider now $i \in \{2, \ldots, N\}$. For the \textit{random} strategy, we randomly sample from the action space, which is a default choice for SAC. For the other strategies, at the beginning of each exploration episode, we choose a previous actor head to generate data instead of the random policy. In the case of the \textit{preceding} strategy, we use the $(i-1)$-th actor's head. For \textit{uniform-previous} policy, we use the $j$-th actor's head, where  $j := \textsc{random\_uniform}(\{1, \ldots, i - 1\})$. Finally, in \textit{best-return} strategy, we first try every possible head, and then act using the $j_{\max}$-th actor's head, where $j_{\max} := \text{argmax}_{j\in \{1, \dots, i-1\}} R^i_j$; $R^i_j$ is the return of the $j$-th head policy on the $i$-th task.

We evaluate how these strategies interact with various CL methods. We pick Fine-tuning, Behavioral cloning, L2, EWC, and PackNet. The results for two well-performing methods, EWC and PackNet, are presented in Table~\ref{table:exploration-main}, with the rest being deferred to Appendix~\ref{sec:experiments_general_cl_appendix}. For EWC, choosing any non-random policy significantly improves upon the baseline random strategy. This is particularly visible in the CW20 sequence, which contains repeated tasks, and arguably can benefit more from an informed strategy like best-return. Interestingly, the results for the rather simple uniform-previous approach are quite competitive. We observe increased performance also for other methods except for PackNet, for which effects are negligible. 

\begin{table}[h!]
  \caption{Average performance and forward transfer for different exploration strategies on CW10 and CW20 sequences. Strategies are added on top of EWC and PackNet methods.}
  \label{table:exploration-main}
  \centering
\begin{tabular}{lll|ll}
\toprule
 Method, exploration   & CW10 perf.              & CW10 f. transfer & CW20 perf.              & CW20 f. transfer  \\
\midrule
 \textbf{EWC, random}               & 0.63 {\tiny [0.60, 0.66]}  & 0.03 {\tiny [-0.04, 0.09]}   & 0.60 {\tiny [0.59, 0.62]}   & -0.14 {\tiny [-0.19, -0.09]}  \\
 \textbf{EWC, preceding}      & \textbf{0.70} {\tiny [0.67, 0.73]}  & 0.09 {\tiny [0.03, 0.15]}   & 0.61 {\tiny [0.59, 0.64]}  & -0.14 {\tiny [-0.19, -0.09]}   \\
 \textbf{EWC, uniform-previous}   & \textbf{0.72} {\tiny [0.69, 0.75]}   & \textbf{0.24} {\tiny [0.19, 0.28]}  & \textbf{0.70} {\tiny [0.68, 0.73]}  & 0.21 {\tiny [0.17, 0.25]}   \\
 \textbf{EWC, best-return}        & \textbf{0.70} {\tiny [0.68, 0.73]}   & \textbf{0.25} {\tiny [0.21, 0.28]}  & \textbf{0.71} {\tiny [0.69, 0.73]}  & \textbf{0.28} {\tiny [0.25, 0.31]}    \\
\midrule
 \textbf{PackNet, random}               & 0.84 {\tiny [0.81, 0.86]}  & 0.26 {\tiny [0.22, 0.29]}    & 0.80 {\tiny [0.79, 0.82]}  & 0.18 {\tiny [0.14, 0.22]}    \\
 \textbf{PackNet, preceding}      & 0.84 {\tiny [0.82, 0.85]}  & 0.24 {\tiny [0.20, 0.27]}    & 0.81 {\tiny [0.80, 0.83]} & 0.20 {\tiny [0.16, 0.24]}    \\
 \textbf{PackNet, uniform-previous}   & 0.84 {\tiny [0.81, 0.86]}  & 0.21 {\tiny [0.15, 0.26]}    & 0.80 {\tiny [0.78, 0.82]}    & 0.23 {\tiny [0.18, 0.27]}    \\
 \textbf{PackNet, best-return}        & 0.85 {\tiny [0.83, 0.86]}  & 0.23 {\tiny [0.20, 0.26]}    & 0.82 {\tiny [0.81, 0.83]}  & 0.23 {\tiny [0.21, 0.25]}    \\
\bottomrule
\end{tabular}
\end{table}

\subsection{Data rehearsal}\label{sec:data_long}
Rehearsal techniques work very well in supervised continual learning \cite{DBLP:journals/corr/abs-1902-10486}. In RL, two major approaches to utilizing previous data have been studied: applying them as offline data using SAC loss, and behavioral cloning of the previous policies. The former, dubbed Perfect memory, was reported to perform poorly \cite{wolczyk2021continual}. Behavioral cloning  achieves more promising results \cite{DBLP:journals/corr/abs-1907-05855,rolnick2019experience}. We study these two approaches with an emphasis on the effects on transfer.

In Perfect memory, all the experiences are kept in the buffer. SAC training is applied to data from the current task and offline data from the previous ones. In Behavioral cloning, an additional small buffer is filled at the end of training on each task. We annotate a subset of samples from the main SAC buffer using the trained actor and critic networks. When training the new task, we sample data from expert buffers and apply auxiliary losses (with different weights), minimizing the KL divergence between current and saved outputs for the actor and L2 distance for the critic; see details in Appendix~\ref{sec:methods_appendix}.

Firstly, we study the effect of rehearsal on transfer in the two-task scenario, using 100 task pairs from CW10, as in Section~\ref{sec:basic_transfer}. We observe that using Perfect memory or cloning both actor and the critic has a detrimental effect on transfer, providing more evidence that critic regularization can be catastrophic. On the other hand, cloning only the actor has a neutral effect; we report results for these and more setups in Appendix~\ref{sec:experiments_general_cl_appendix}. As such, in the remaining Behavioral cloning experiments, we regularize only the actor, unless noted otherwise.

\begin{table}[h!]
  \caption{Average performance and forward transfer for Perfect memory and Behavioral cloning methods, as described in Section~\ref{sec:data_long}. Fine-tuning and PackNet are included for reference.}
  \label{table:er-cw10}
  \centering
\begin{tabular}{lll|ll}
\toprule
 method       & CW10 perf.         & CW10 f. transfer  & CW20 perf.         & CW20 f. transfer  \\
\midrule
 \textbf{Perfect memory}       & 0.27 {\tiny [0.24, 0.30]}   & -1.13 {\tiny [-1.23, -1.04]}  & 0.09 {\tiny [0.06, 0.12]}   & -1.32 {\tiny [-1.41, -1.24]}  \\

 \textbf{Behavioral cloning} & \textbf{0.84} {\tiny [0.81, 0.86]}   & \textbf{0.41} {\tiny [0.38, 0.43]}   & \textbf{0.83} {\tiny [0.81, 0.85]}   & \textbf{0.36} {\tiny [0.34, 0.38]}    \\
 
  \textbf{Fine-tuning}            & 0.10 {\tiny [0.10, 0.10]}   & 0.31 {\tiny [0.27, 0.34]}    & 0.05 {\tiny [0.05, 0.05]}   & 0.19 {\tiny [0.15, 0.23]}     \\

 \textbf{PackNet}         & \textbf{0.84} {\tiny [0.81, 0.86]}  & 0.26 {\tiny [0.22, 0.29]}   & 0.80 {\tiny [0.79, 0.82]}  & 0.18 {\tiny [0.14, 0.22]}    \\
 
\bottomrule
\end{tabular}
\end{table}

Secondly, we perform experiments on longer sequences, CW10 and CW20; see Table~\ref{table:er-cw10}. For reference, we include two methods tested in \cite{wolczyk2021continual}, Fine-tuning and PackNet. Fine-tuning achieves the highest transfer and PackNet the highest overall performance out of the methods tested in \cite{wolczyk2021continual}. Behavioral cloning performs very well. In terms of the average performance, it is on par with PackNet on CW10 and better on CW20. Importantly, it significantly outperforms the baselines in terms of transfer. We can see that Perfect memory works poorly, in line with the existing literature. In Appendix~\ref{sec:results_appendix}, we present the results for five other CL methods benchmarked in \cite{wolczyk2021continual}. 

In the end, we observe an interesting phenomenon. While behavioral cloning does not improve transfer in two-task scenario, it has a positive effect for the longer sequences. This result hints that perhaps  the learner accumulates knowledge of the previous tasks and, thus, can reuse the most relevant parts of the past to improve the training of the current task. Additionally, perhaps behavioral cloning loss acts as a regularizer and helps shape more general features, thus further improving transfer.

\subsection{Regularizing the critic}
\label{sec:regularizing_critic}
This section is devoted to the study of critic regularization in CRL methods. Since in our formulation of the problem, the primary objective of CRL is the final performance of the actor, we have some flexibility in how we treat the critic. We can even completely ignore forgetting in the critic, as recommended in \cite{wolczyk2021continual}. Other works suggest that regularization might be beneficial \cite{DBLP:journals/corr/abs-2106-02940}.

To understand this issue better, we carefully measure the performance while varying the strength of the regularization, by changing the critic regularization coefficients for EWC, L2, and Behavioral cloning. 
We first find a good value for the actor regularization coefficient, with the critic regularization coefficient being set to $0$. Then, with this value, we perform the sweep over the critic coefficients, covering a wide range from $\num{1e-10}$ to $100$, and run training on the CW10 sequence. For all three methods, we observe that for the smallest values of critic regularization, the performance is similar to the version without critic regularization, and then after some threshold, performance visibly deteriorates. In the case of Behavioral cloning, it drops from $0.82$ (no critic regularization) to $0.77$ (critic regularization coefficient $= 0.001$) and then further, see Table~\ref{table:cw10-bc-critic-reg}. The complete results are presented in Appendix~\ref{app:reg-critic}.

This confirms the practical recommendation from \cite{wolczyk2021continual} to regularize only the actor. One possible explanation is that $TD$-learning used for the critic is highly sensitive to biases introduced by regularization.

\section{Combining the improvements: \method{}}\label{sec:method}
Based on the experimental findings presented so far, we propose to combine the discovered enhancements in a simple method for continual reinforcement learning. This method significantly improves the performance in the Continual World benchmark \cite{wolczyk2021continual}. In particular, we observe a sharp transfer increase to a value that matches a soft upper bound for transfer introduced in \cite{wolczyk2021continual}.

We incorporate the following choices in the proposed method:
\begin{itemize}
    \item We use behavioral cloning for the actor, which, as we showed in Section~\ref{sec:data_long}, effectively mitigates forgetting and increases transfer.
    \item We use best-return exploration, as described in Section~\ref{sec:exploration}, which efficiently reuses old policy heads for faster exploration.
    \item As indicated in Section~\ref{sec:regularizing_critic}, we do not use any CL regularization for the critic.
    \item We use multiple output heads for both actor and critic to profit from transferred representations without introducing too much bias in the new tasks, as discussed in Section \ref{sec:basic_transfer}.
\end{itemize}

We dub the method \method{} to reflect the usage of the behavioral \textbf{clon}ing, improved \textbf{ex}ploration, and SAC algorithm. 

We compare \method{} with the behavioral cloning and  $7$ methods considered in \cite{wolczyk2021continual}, on the CW10 and CW20 sequences. We present results in Figure~\ref{fig:main_clonex_results} (see Introduction) and Appendix~\ref{sec:results_appendix}. \method{} achieves $87\%$ final performance compared to $80\%$ of PackNet, the best method in \cite{wolczyk2021continual}.

The forward transfer of \method{}, improves sharply from $0.19$, the best previous result, to $0.54$. Notably, \method{}'s result closely matches the reference forward transfer, see below. We conjecture that this excellent transfer is an important factor in the final performance. We also notice that improvements brought separately by behavioral cloning and the best-return exploration strategy work well together.

\textbf{Reference forward transfer} (RT) was introduced in \cite{wolczyk2021continual} as a soft upper bound for transfer. For a sequence of tasks $t_1, \ldots, t_N$, it is defined defined as $\text{RT}:=\frac{1}{N}\sum_{i=2}^N \max_{j<i} \text{FT}(t_j, t_i)$, where $F(t_j, t_i)$ denotes the forward transfer for the pair of tasks $t_j, t_i$.

Intuitively, $RT$ estimates the level of forward transfer, which could be achieved when a method is able to remember and transfer all meaningful aspects of previously seen tasks. Note that in principle, higher values of $RT$ could still be achievable if the knowledge from the previous tasks is composed. In our setup, the values of $RT$ are $0.44$ for CW10 and $0.55$ for CW20. In both cases, they are closely matched by the forward transfer of \method{}. We note that our $RT$ values are higher than the one reported in \cite{wolczyk2021continual}, since their work does not take into account the effects of improved exploration.

\section{Limitations} \label{sec:limitations} 
We are fully aware that our analyses do not cover the entire spectrum of problems that one might be interested in when studying transfer in CRL. 
Here, we summarize a few limitations of our work:

\begin{itemize}
  \item We build on top of the SAC algorithm. There is a risk that some of the conclusions from this paper would differ for another choice of the underlying RL method.
  \item We focus on the Continual World suite. There is a possibility that some of the results from this paper would differ in environments from other domains or with different, potentially structured state spaces. 
  \item \method{} requires retaining data from previous tasks, which may not always be feasible (e.g., due to privacy concerns).
\end{itemize}

\section{Conclusions}
In this work, we identify  
and study some of the key factors contributing to transfer in continual reinforcement learning.
In the first part of the study, we focus on the transfer alone, disregarding other CL desiderata, and analyze how different components of the SAC algorithm (actor, critic, exploration) contribute to it. We identify the critic as the leading component.
 
In the second part, we study further effects that are relevant to the full continual learning setup with long task sequences. In particular, we show that behavioral cloning and reusing previous policies for exploration significantly improve both transfer and the final performance. This leads to a new method, \method{}, which outperforms considered baselines. 

We believe that this work constitutes the first step toward understanding the mechanisms behind transfer in continual reinforcement learning. There are still important issues to be resolved, e.g., pinpointing the exact role of feature reuse or the interplay between transfer and forgetting. We hope that these will be addressed by the community in the future.

\begin{ack}
The work of Maciej Wołczyk was supported by the National Centre of Science (Poland) Grant No. 2021/43/B/ST6/01456. The work of Piotr Miłoś was supported by the Polish National Science Center grant UMO-2017/26/E/ST6/00622 and UMO-2019/35/O/ST6/03464. This research was supported by the PL-Grid Infrastructure. Our experiments were managed using \url{https://neptune.ai}. We would like to thank the Neptune team for providing us access to the team version and technical support.
\end{ack}

\bibliography{bibliography}

\begin{thebibliography}{10}

\bibitem{SpinningUp2018}
Joshua Achiam.
\newblock {Spinning Up in Deep Reinforcement Learning}.
\newblock 2018.

\bibitem{DBLP:conf/eccv/AljundiBERT18}
Rahaf Aljundi, Francesca Babiloni, Mohamed Elhoseiny, Marcus Rohrbach, and
  Tinne Tuytelaars.
\newblock Memory aware synapses: Learning what (not) to forget.
\newblock In Vittorio Ferrari, Martial Hebert, Cristian Sminchisescu, and Yair
  Weiss, editors, {\em Computer Vision - {ECCV} 2018 - 15th European
  Conference, Munich, Germany, September 8-14, 2018, Proceedings, Part {III}},
  volume 11207 of {\em Lecture Notes in Computer Science}, pages 144--161.
  Springer, 2018.

\bibitem{andreas2017modular}
Jacob Andreas, Dan Klein, and Sergey Levine.
\newblock Modular multitask reinforcement learning with policy sketches.
\newblock In {\em International Conference on Machine Learning}, pages
  166--175. PMLR, 2017.

\bibitem{DBLP:journals/corr/abs-1909-07528}
Bowen Baker, Ingmar Kanitscheider, Todor~M. Markov, Yi~Wu, Glenn Powell, Bob
  McGrew, and Igor Mordatch.
\newblock Emergent tool use from multi-agent autocurricula.
\newblock {\em CoRR}, abs/1909.07528, 2019.

\bibitem{brys2015policy}
Tim Brys, Anna Harutyunyan, Matthew~E Taylor, and Ann Now{\'e}.
\newblock Policy transfer using reward shaping.
\newblock In {\em AAMAS}, pages 181--188, 2015.

\bibitem{DBLP:conf/iclr/ChaudhryRRE19}
Arslan Chaudhry, Marc'Aurelio Ranzato, Marcus Rohrbach, and Mohamed Elhoseiny.
\newblock Efficient lifelong learning with {A-GEM}.
\newblock In {\em 7th International Conference on Learning Representations,
  {ICLR} 2019, New Orleans, LA, USA, May 6-9, 2019}. OpenReview.net, 2019.

\bibitem{DBLP:journals/corr/abs-1902-10486}
Arslan Chaudhry, Marcus Rohrbach, Mohamed Elhoseiny, Thalaiyasingam Ajanthan,
  Puneet~Kumar Dokania, Philip H.~S. Torr, and Marc'Aurelio Ranzato.
\newblock Continual learning with tiny episodic memories.
\newblock {\em CoRR}, abs/1902.10486, 2019.

\bibitem{de2019continual}
Matthias De~Lange, Rahaf Aljundi, Marc Masana, Sarah Parisot, Xu~Jia, Ales
  Leonardis, Gregory Slabaugh, and Tinne Tuytelaars.
\newblock A continual learning survey: Defying forgetting in classification
  tasks.
\newblock {\em arXiv preprint arXiv:1909.08383}, 2019.

\bibitem{devin2017learning}
Coline Devin, Abhishek Gupta, Trevor Darrell, Pieter Abbeel, and Sergey Levine.
\newblock Learning modular neural network policies for multi-task and
  multi-robot transfer.
\newblock In {\em 2017 IEEE international conference on robotics and automation
  (ICRA)}, pages 2169--2176. IEEE, 2017.

\bibitem{bert}
Jacob Devlin, Ming-Wei Chang, Kenton Lee, and Kristina~N. Toutanova.
\newblock Bert: Pre-training of deep bidirectional transformers for language
  understanding.
\newblock 2018.

\bibitem{dosovitskiy2020image}
Alexey Dosovitskiy, Lucas Beyer, Alexander Kolesnikov, Dirk Weissenborn,
  Xiaohua Zhai, Thomas Unterthiner, Mostafa Dehghani, Matthias Minderer, Georg
  Heigold, Sylvain Gelly, et~al.
\newblock An image is worth 16x16 words: Transformers for image recognition at
  scale.
\newblock {\em arXiv preprint arXiv:2010.11929}, 2020.

\bibitem{DBLP:journals/corr/abs-1805-09733}
Sebastian Farquhar and Yarin Gal.
\newblock Towards robust evaluations of continual learning.
\newblock {\em CoRR}, abs/1805.09733, 2018.

\bibitem{fernando2017pathnet}
Chrisantha Fernando, Dylan Banarse, Charles Blundell, Yori Zwols, David Ha,
  Andrei~A Rusu, Alexander Pritzel, and Daan Wierstra.
\newblock Pathnet: Evolution channels gradient descent in super neural
  networks.
\newblock {\em arXiv preprint arXiv:1701.08734}, 2017.

\bibitem{fujimoto2022i}
Scott Fujimoto, David Meger, Doina Precup, Ofir Nachum, and Shixiang~Shane Gu.
\newblock Why should i trust you, bellman? the bellman error is a poor
  replacement for value error, 2022.

\bibitem{haarnoja2018soft}
Tuomas Haarnoja, Aurick Zhou, Pieter Abbeel, and Sergey Levine.
\newblock Soft actor-critic: Off-policy maximum entropy deep reinforcement
  learning with a stochastic actor.
\newblock In {\em International conference on machine learning}, pages
  1861--1870. PMLR, 2018.

\bibitem{DBLP:journals/corr/abs-1812-05905}
Tuomas Haarnoja, Aurick Zhou, Kristian Hartikainen, George Tucker, Sehoon Ha,
  Jie Tan, Vikash Kumar, Henry Zhu, Abhishek Gupta, Pieter Abbeel, and Sergey
  Levine.
\newblock Soft actor-critic algorithms and applications.
\newblock {\em CoRR}, abs/1812.05905, 2018.

\bibitem{HADSELL20201028}
Raia Hadsell, Dushyant Rao, Andrei~A. Rusu, and Razvan Pascanu.
\newblock Embracing change: Continual learning in deep neural networks.
\newblock {\em Trends in Cognitive Sciences}, 24(12):1028 -- 1040, 2020.

\bibitem{he2016deep}
Kaiming He, Xiangyu Zhang, Shaoqing Ren, and Jian Sun.
\newblock Deep residual learning for image recognition.
\newblock In {\em Proceedings of the IEEE conference on computer vision and
  pattern recognition}, pages 770--778, 2016.

\bibitem{huang2021continual}
Yizhou Huang, Kevin Xie, Homanga Bharadhwaj, and Florian Shkurti.
\newblock Continual model-based reinforcement learning with hypernetworks.
\newblock In {\em 2021 IEEE International Conference on Robotics and Automation
  (ICRA)}, pages 799--805. IEEE, 2021.

\bibitem{kaplanis2018continual}
Christos Kaplanis, Murray Shanahan, and Claudia Clopath.
\newblock Continual reinforcement learning with complex synapses.
\newblock In {\em International Conference on Machine Learning}, pages
  2497--2506. PMLR, 2018.

\bibitem{DBLP:journals/corr/abs-2106-02940}
Samuel Kessler, Jack Parker{-}Holder, Philip~J. Ball, Stefan Zohren, and
  Stephen~J. Roberts.
\newblock Same state, different task: Continual reinforcement learning without
  interference.
\newblock {\em CoRR}, abs/2106.02940, 2021.

\bibitem{khetarpal2020continual}
Khimya Khetarpal, Matthew Riemer, Irina Rish, and Doina Precup.
\newblock Towards continual reinforcement learning: A review and perspectives,
  2020.

\bibitem{DBLP:journals/corr/KirkpatrickPRVD16}
James Kirkpatrick, Razvan Pascanu, Neil~C. Rabinowitz, Joel Veness, Guillaume
  Desjardins, Andrei~A. Rusu, Kieran Milan, John Quan, Tiago Ramalho, Agnieszka
  Grabska{-}Barwinska, Demis Hassabis, Claudia Clopath, Dharshan Kumaran, and
  Raia Hadsell.
\newblock Overcoming catastrophic forgetting in neural networks.
\newblock {\em CoRR}, abs/1612.00796, 2016.

\bibitem{DBLP:conf/nips/KrizhevskySH12}
Alex Krizhevsky, Ilya Sutskever, and Geoffrey~E. Hinton.
\newblock Imagenet classification with deep convolutional neural networks.
\newblock In Peter~L. Bartlett, Fernando C.~N. Pereira, Christopher J.~C.
  Burges, L{\'{e}}on Bottou, and Kilian~Q. Weinberger, editors, {\em Advances
  in Neural Information Processing Systems 25: 26th Annual Conference on Neural
  Information Processing Systems 2012. Proceedings of a meeting held December
  3-6, 2012, Lake Tahoe, Nevada, United States}, pages 1106--1114, 2012.

\bibitem{lu2021pretrained}
Kevin Lu, Aditya Grover, Pieter Abbeel, and Igor Mordatch.
\newblock Pretrained transformers as universal computation engines.
\newblock {\em arXiv preprint arXiv:2103.05247}, 2021.

\bibitem{DBLP:conf/cvpr/MallyaL18}
Arun Mallya and Svetlana Lazebnik.
\newblock Packnet: Adding multiple tasks to a single network by iterative
  pruning.
\newblock In {\em 2018 {IEEE} Conference on Computer Vision and Pattern
  Recognition, {CVPR} 2018, Salt Lake City, UT, USA, June 18-22, 2018}, pages
  7765--7773. {IEEE} Computer Society, 2018.

\bibitem{mehta2008transfer}
Neville Mehta, Sriraam Natarajan, Prasad Tadepalli, and Alan Fern.
\newblock Transfer in variable-reward hierarchical reinforcement learning.
\newblock {\em Machine Learning}, 73(3):289--312, 2008.

\bibitem{neyshabur2020being}
Behnam Neyshabur, Hanie Sedghi, and Chiyuan Zhang.
\newblock What is being transferred in transfer learning?
\newblock {\em Advances in neural information processing systems}, 33:512--523,
  2020.

\bibitem{NEURIPS2020_0607f4c7}
Behnam Neyshabur, Hanie Sedghi, and Chiyuan Zhang.
\newblock What is being transferred in transfer learning?
\newblock In H.~Larochelle, M.~Ranzato, R.~Hadsell, M.F. Balcan, and H.~Lin,
  editors, {\em Advances in Neural Information Processing Systems}, volume~33,
  pages 512--523. Curran Associates, Inc., 2020.

\bibitem{ng1999policy}
Andrew~Y Ng, Daishi Harada, and Stuart Russell.
\newblock Policy invariance under reward transformations: Theory and
  application to reward shaping.
\newblock In {\em Icml}, volume~99, pages 278--287, 1999.

\bibitem{DBLP:conf/iclr/NguyenLBT18}
Cuong~V. Nguyen, Yingzhen Li, Thang~D. Bui, and Richard~E. Turner.
\newblock Variational continual learning.
\newblock In {\em 6th International Conference on Learning Representations,
  {ICLR} 2018, Vancouver, BC, Canada, April 30 - May 3, 2018, Conference Track
  Proceedings}. OpenReview.net, 2018.

\bibitem{DBLP:journals/nn/ParisiKPKW19}
German~Ignacio Parisi, Ronald Kemker, Jose~L. Part, Christopher Kanan, and
  Stefan Wermter.
\newblock Continual lifelong learning with neural networks: {A} review.
\newblock {\em Neural Networks}, 113:54--71, 2019.

\bibitem{raghu2019rapid}
Aniruddh Raghu, Maithra Raghu, Samy Bengio, and Oriol Vinyals.
\newblock Rapid learning or feature reuse? towards understanding the
  effectiveness of maml.
\newblock {\em arXiv preprint arXiv:1909.09157}, 2019.

\bibitem{DBLP:journals/corr/abs-2201-12122}
Machel Reid, Yutaro Yamada, and Shixiang~Shane Gu.
\newblock Can wikipedia help offline reinforcement learning?
\newblock {\em CoRR}, abs/2201.12122, 2022.

\bibitem{rolnick2019experience}
David Rolnick, Arun Ahuja, Jonathan Schwarz, Timothy Lillicrap, and Gregory
  Wayne.
\newblock Experience replay for continual learning.
\newblock {\em Advances in Neural Information Processing Systems}, 32, 2019.

\bibitem{rusu2016progressive}
Andrei~A Rusu, Neil~C Rabinowitz, Guillaume Desjardins, Hubert Soyer, James
  Kirkpatrick, Koray Kavukcuoglu, Razvan Pascanu, and Raia Hadsell.
\newblock Progressive neural networks.
\newblock {\em arXiv preprint arXiv:1606.04671}, 2016.

\bibitem{shao2019survey}
Kun Shao, Zhentao Tang, Yuanheng Zhu, Nannan Li, and Dongbin Zhao.
\newblock A survey of deep reinforcement learning in video games.
\newblock {\em arXiv preprint arXiv:1912.10944}, 2019.

\bibitem{steccanella2020hierarchical}
Lorenzo Steccanella, Simone Totaro, Damien Allonsius, and Anders Jonsson.
\newblock Hierarchical reinforcement learning for efficient exploration and
  transfer.
\newblock {\em arXiv preprint arXiv:2011.06335}, 2020.

\bibitem{talvitie2007experts}
Erik Talvitie and Satinder~P Singh.
\newblock An experts algorithm for transfer learning.
\newblock In {\em IJCAI}, pages 1065--1070, 2007.

\bibitem{10.1007/978-3-030-01424-7_27}
Chuanqi Tan, Fuchun Sun, Tao Kong, Wenchang Zhang, Chao Yang, and Chunfang Liu.
\newblock A survey on deep transfer learning.
\newblock In V{\v{e}}ra K{\r{u}}rkov{\'a}, Yannis Manolopoulos, Barbara Hammer,
  Lazaros Iliadis, and Ilias Maglogiannis, editors, {\em Artificial Neural
  Networks and Machine Learning -- ICANN 2018}, pages 270--279, Cham, 2018.
  Springer International Publishing.

\bibitem{taylor2009transfer}
Matthew~E Taylor and Peter Stone.
\newblock Transfer learning for reinforcement learning domains: A survey.
\newblock {\em Journal of Machine Learning Research}, 10(7), 2009.

\bibitem{taylor2007transfer}
Matthew~E Taylor, Peter Stone, and Yaxin Liu.
\newblock Transfer learning via inter-task mappings for temporal difference
  learning.
\newblock {\em Journal of Machine Learning Research}, 8(9), 2007.

\bibitem{DBLP:journals/corr/abs-2107-12808}
Open Ended~Learning Team, Adam Stooke, Anuj Mahajan, Catarina Barros, Charlie
  Deck, Jakob Bauer, Jakub Sygnowski, Maja Trebacz, Max Jaderberg,
  Micha{\"{e}}l Mathieu, Nat McAleese, Nathalie Bradley{-}Schmieg, Nathaniel
  Wong, Nicolas Porcel, Roberta Raileanu, Steph Hughes{-}Fitt, Valentin
  Dalibard, and Wojciech~Marian Czarnecki.
\newblock Open-ended learning leads to generally capable agents.
\newblock {\em CoRR}, abs/2107.12808, 2021.

\bibitem{DBLP:journals/corr/abs-1907-05855}
Ren{\'{e}} Traor{\'{e}}, Hugo Caselles{-}Dupr{\'{e}}, Timoth{\'{e}}e Lesort,
  Te~Sun, Guanghang Cai, Natalia~D{\'{\i}}az Rodr{\'{\i}}guez, and David
  Filliat.
\newblock Discorl: Continual reinforcement learning via policy distillation.
\newblock {\em CoRR}, abs/1907.05855, 2019.

\bibitem{DBLP:conf/nips/VaswaniSPUJGKP17}
Ashish Vaswani, Noam Shazeer, Niki Parmar, Jakob Uszkoreit, Llion Jones,
  Aidan~N. Gomez, Lukasz Kaiser, and Illia Polosukhin.
\newblock Attention is all you need.
\newblock In Isabelle Guyon, Ulrike von Luxburg, Samy Bengio, Hanna~M. Wallach,
  Rob Fergus, S.~V.~N. Vishwanathan, and Roman Garnett, editors, {\em Advances
  in Neural Information Processing Systems 30: Annual Conference on Neural
  Information Processing Systems 2017, December 4-9, 2017, Long Beach, CA,
  {USA}}, pages 5998--6008, 2017.

\bibitem{wolczyk2021continual}
Maciej Wo{\l}czyk, Micha{\l} Zaj{\k{a}}c, Razvan Pascanu, Lukasz Kucinski, and
  Piotr Mi{\l}o{\'s}.
\newblock Continual world: A robotic benchmark for continual reinforcement
  learning.
\newblock {\em Advances in Neural Information Processing Systems}, 34, 2021.

\bibitem{yu2020gradient}
Tianhe Yu, Saurabh Kumar, Abhishek Gupta, Sergey Levine, Karol Hausman, and
  Chelsea Finn.
\newblock Gradient surgery for multi-task learning.
\newblock {\em Advances in Neural Information Processing Systems},
  33:5824--5836, 2020.

\bibitem{metaworld}
Tianhe Yu, Deirdre Quillen, Zhanpeng He, Ryan Julian, Karol Hausman, Chelsea
  Finn, and Sergey Levine.
\newblock Meta-world: {A} benchmark and evaluation for multi-task and meta
  reinforcement learning.
\newblock In Leslie~Pack Kaelbling, Danica Kragic, and Komei Sugiura, editors,
  {\em 3rd Annual Conference on Robot Learning, CoRL 2019, Osaka, Japan,
  October 30 - November 1, 2019, Proceedings}, volume 100 of {\em Proceedings
  of Machine Learning Research}, pages 1094--1100. {PMLR}, 2019.

\bibitem{DBLP:journals/corr/abs-1910-04867}
Xiaohua Zhai, Joan Puigcerver, Alexander Kolesnikov, Pierre Ruyssen, Carlos
  Riquelme, Mario Lucic, Josip Djolonga, Andr{\'{e}}~Susano Pinto, Maxim
  Neumann, Alexey Dosovitskiy, Lucas Beyer, Olivier Bachem, Michael Tschannen,
  Marcin Michalski, Olivier Bousquet, Sylvain Gelly, and Neil Houlsby.
\newblock The visual task adaptation benchmark.
\newblock {\em CoRR}, abs/1910.04867, 2019.

\bibitem{zhu2020transfer}
Zhuangdi Zhu, Kaixiang Lin, and Jiayu Zhou.
\newblock Transfer learning in deep reinforcement learning: A survey.
\newblock {\em arXiv preprint arXiv:2009.07888}, 2020.

\bibitem{zhuang2019comprehensive}
Fuzhen Zhuang, Zhiyuan Qi, Keyu Duan, Dongbo Xi, Yongchun Zhu, Hengshu Zhu, Hui
  Xiong, and Qing He.
\newblock A comprehensive survey on transfer learning, 2021.

\end{thebibliography}
\bibliographystyle{plain}
\section*{Checklist}

The checklist follows the references.  Please
read the checklist guidelines carefully for information on how to answer these
questions.  For each question, change the default \answerTODO{} to \answerYes{},
\answerNo{}, or \answerNA{}.  You are strongly encouraged to include a {\bf
justification to your answer}, either by referencing the appropriate section of
your paper or providing a brief inline description.  For example:
\begin{itemize}
  \item Did you include the license to the code and datasets? \answerYes{See Section~??.}
  \item Did you include the license to the code and datasets? \answerNo{The code and the data are proprietary.}
  \item Did you include the license to the code and datasets? \answerNA{}
\end{itemize}
Please do not modify the questions and only use the provided macros for your
answers.  Note that the Checklist section does not count towards the page
limit.  In your paper, please delete this instructions block and only keep the
Checklist section heading above along with the questions/answers below.

\begin{enumerate}

\item For all authors...
\begin{enumerate}
  \item Do the main claims made in the abstract and introduction accurately reflect the paper's contributions and scope?
    \answerYes{}
  \item Did you describe the limitations of your work?
    \answerYes{In Section \ref{sec:limitations}}
  \item Did you discuss any potential negative societal impacts of your work?
    \answerNo{We do not see this work as having a significant societal impact.}
  \item Have you read the ethics review guidelines and ensured that your paper conforms to them?
    \answerYes{}
\end{enumerate}

\item If you are including theoretical results...
\begin{enumerate}
  \item Did you state the full set of assumptions of all theoretical results?
    \answerNA{}
        \item Did you include complete proofs of all theoretical results?
    \answerNA{}
\end{enumerate}

\item If you ran experiments...
\begin{enumerate}
  \item Did you include the code, data, and instructions needed to reproduce the main experimental results (either in the supplemental material or as a URL)?
    \answerYes{The code, including the scripts used to run the experiments from the paper, are in the supplementary materials.}
  \item Did you specify all the training details (e.g., data splits, hyperparameters, how they were chosen)?
    \answerYes{}
        \item Did you report error bars (e.g., with respect to the random seed after running experiments multiple times)?
    \answerYes{We conduct each experiment with multiple seeds (at least 10)}
        \item Did you include the total amount of compute and the type of resources used (e.g., type of GPUs, internal cluster, or cloud provider)?
    \answerYes{We describe these details in Appendix \ref{sec:infra_appendix}.}
\end{enumerate}

\item If you are using existing assets (e.g., code, data, models) or curating/releasing new assets...
\begin{enumerate}
  \item If your work uses existing assets, did you cite the creators?
    \answerNA{}
  \item Did you mention the license of the assets?
    \answerNA{}
  \item Did you include any new assets either in the supplemental material or as a URL?
    \answerNA{}
  \item Did you discuss whether and how consent was obtained from people whose data you're using/curating?
    \answerNA{}
  \item Did you discuss whether the data you are using/curating contains personally identifiable information or offensive content?
    \answerNA{}
\end{enumerate}

\item If you used crowdsourcing or conducted research with human subjects...
\begin{enumerate}
  \item Did you include the full text of instructions given to participants and screenshots, if applicable?
    \answerNA{}
  \item Did you describe any potential participant risks, with links to Institutional Review Board (IRB) approvals, if applicable?
    \answerNA{}
  \item Did you include the estimated hourly wage paid to participants and the total amount spent on participant compensation?
    \answerNA{}
\end{enumerate}

\end{enumerate}


\newpage

\appendix

\section{Technical details}
\label{sec:setting_appendix}

\subsection{Hyperparameters}
\label{sec:hyperparams_appendix}

\begin{table}[h!]
    \centering
    \begin{tabular}{ lc }
        \toprule
        parameter                    & value \\
        \midrule
        learning rate                &  $\num{1e-3}$   \\
        batch size                   & $128$          \\
        discount factor $\gamma$     & $0.99$         \\
        target output std $\sigma_t$ & $0.089$        \\
        replay buffer size           & 1M             \\
        \bottomrule
    \end{tabular}
    \caption{Hyperparameters used for the underlying SAC algorithm.
    }
    \label{table:hparams-general}
\end{table}

We use 
hyperparameters for 
the underlying SAC algorithm as in \cite{wolczyk2021continual}; for completeness we present them in Table~\ref{table:hparams-general}. We use the Adam optimizer with the default values of $\beta_1 = 0.9, \beta_2 = 0.999, \epsilon = 10^{-8}$.

Below we list tested hyperparameter ranges and final values for individual CL methods; for baselines included in Continual World, ranges were based on the optimal value reported in \cite{wolczyk2021continual}:
\begin{itemize}
    \item L2: we search regularization parameter in $\{1000, 10000, 100000, 1000000\}$. Selected value is $100000$.
    \item EWC: we search regularization parameter in $\{1000, 10000, 100000, 1000000\}$. Selected value is $10000$.
    \item MAS: we search regularization parameter in $\{1000, 10000, 100000, 1000000\}$. Selected value is $1000$.
    \item VCL: we search regularization parameter in $\{0.01, 0.1, 1, 10\}$. Selected value is $1$.
    \item PackNet: we use the number of retraining steps $= 100000$, and we use global gradient norm clipping \num{2e-5}, as in \cite{wolczyk2021continual}.
    \item Perfect Memory: we search over batch size in $\{128, 256, 512\}$. Selected value is $256$.
    \item A-GEM: we search over episodic batch size in $\{128, 256\}$. Selected value is $128$.
    \item Behavioral cloning: we search over actor's regularization coefficient in $\{10, 100, 1000\}$, episodic batch size in $\{128, 256\}$; selected values are $(100, 128)$. Episodic memory per task is set to $M=10000$, and we use global gradient norm clipping $0.1$ for stability. After we tuned these parameters, we searched regularization coefficient for the critic from values $\{ 0, \num{1e-4}, \num{1e-3}, 0.01, 0.1, 1, 10, 100 \}$; the selected value is $0$ (no regularization for critic).
    
\end{itemize}

\subsection{SAC}
\label{sec:sac_appendix}

In this work, we use SAC \cite{haarnoja2018soft} as an underlying RL algorithm. It is an off-policy actor-critic algorithm, based on the maximum entropy principle. The critic approximates the entropy-adjusted $Q$-function under the current policy and is trained with the following loss \cite{SpinningUp2018}:

$$ \mathbb{E}_{(s,a,r,s',d)} \left[ (Q_{\phi_i}(s, a) - \hat{Q})^2 \right],$$
where
$$\hat{Q} := r + \gamma (1-d) \left ( \min_{j=1,2} Q_{\phi_{target,j}}(s', a') - \alpha \log (\pi_{\theta}(a'|s') \right ) , \quad
a' \sim \pi_{\theta}(\cdot | s'). $$

The actor searches for actions that maximize the $Q$-function, i.e.\ it maximizes the following objective:
$$ \mathbb{E}_{s, a \sim \pi} \left [ Q^{\pi}(s, a) - \alpha \log \pi(a | s) \right ]. $$

\section{Details on the methods}
\label{sec:methods_appendix}

\subsection{Behavioral cloning} In Behavioral cloning
(see Algorithm \ref{alg:bc}), at the end of each task, we randomly sample a subset from the SAC buffer, label it using the outputs of the current (trained) networks and add it to a separate buffer as "expert" data. In the subsequent tasks, we add auxiliary losses to the SAC's objective to 
imitate these expert data; for the actor, we use the KL divergence, and for the critics, we use the L2 loss (which can be derived as KL divergence between mean-parameterized Gaussian distributions). 

\subsection{Baselines from Continual World benchmark}
Here, we briefly describe the baselines from Continual World that we are using, and refer the reader to Appendix B of \cite{wolczyk2021continual} for more details. The simplest continual learning  baseline is \textbf{Fine-tuning} where the model is simply trained on the sequence of tasks without applying any kind of mechanism for avoiding forgetting or encouraging forward transfer. Then, we consider three standard methods from the family of regularization methods, \textbf{L2} \cite{DBLP:journals/corr/KirkpatrickPRVD16}, \textbf{EWC} \cite{DBLP:journals/corr/KirkpatrickPRVD16} and \textbf{MAS} \cite{DBLP:conf/eccv/AljundiBERT18} which apply quadratic regularization to network weights while using different mechanisms for establishing per-parameter regularization coefficients. \textbf{VCL} \cite{DBLP:conf/iclr/NguyenLBT18} applies variational inference to Bayesian neural networks to facilitate continual learning. Using a different approach, \textbf{A-GEM} \cite{DBLP:conf/iclr/ChaudhryRRE19} projects the gradients according to constraints obtained from data from the buffer. \textbf{Perfect memory} is a simple method that keeps all of the data from the past tasks in the SAC's buffer. Finally, \textbf{PackNet} \cite{DBLP:conf/cvpr/MallyaL18} freezes a fraction of the parameters of the network after each task so that the performance does not deteriorate on the previous tasks.

\DeclarePairedDelimiterX{\infdivx}[2]{(}{)}{%
  #1\;\delimsize\|\;#2%
}
\newcommand{\infdiv}{D_{KL}\infdivx}
\DeclarePairedDelimiter{\norm}{\lVert}{\rVert}

\begin{algorithm}[t]
\caption{Behavioral cloning}
\label{alg:bc}
\begin{algorithmic}[1]
\State \textbf{input:} number of tasks $N$, SAC actor $\pi$, SAC critics $q_1, q_2$, expert buffer $\mathcal{D}_{ex} := \emptyset$, expert batch size $B_{ex}$, regularization coefficients for actor and critic $r_{actor}, r_{critic}$
\State Train SAC on task $t_1$.
\State Gather actor and critic outputs as targets to populate $\mathcal{D}_{ex}$.

\For{task $t_i$, $i := 2, \ldots, N$}
\State Train SAC on task $t_i$, with the following modified update rule:
\State \hspace{\algorithmicindent} Compute the SAC loss $l_{SAC}$
\State \hspace{\algorithmicindent} Sample $(o^{1..B_{ex}}, \hat{\pi}^{1..B_{ex}}, \hat{q}_1^{1..B_{ex}}, \hat{q}_2^{1..B_{ex}}) \sim \mathcal{D}_{ex}$  
\Comment{state, tgt policy dist, tgt  preds}

\State \hspace{\algorithmicindent} $l_{actor} := \frac{1}{B_{ex}}
\sum_{j=1}^{B_{ex}} \infdiv{\pi(o^j)}{\hat{\pi}^j}$
\State \hspace{\algorithmicindent} $l_{critic} := \frac{1}{B_{ex}} \sum_{j=1}^{B_{ex}} \left ((q_1(o^j) - \hat{q}_1^j)^2 + (q_2(o^j) - \hat{q}_2^j)^2 \right ).$
\State \hspace{\algorithmicindent} Minimize $l_\text{SAC} + r_{actor} \cdot l_{actor} + r_{critic} \cdot l_{critic} $

\State Gather actor and critic outputs as targets to extend $\mathcal{D}_{ex}$.
\EndFor
\end{algorithmic}
\end{algorithm}

 \section{Results for all methods}
 \label{sec:results_appendix}
 Results for all methods, including the baselines from Continual World, as well as \method{} and its ablations, are presented in Table~\ref{table:all-methods-cw10-full} (CW10) and Table~\ref{table:all-methods-cw20-full} (CW20).
 
 \begin{table}[h!]
   \caption{Results of all the methods on CW10 sequence. Average performance, forgetting, and forward transfer are shown in columns. 90\% bootstrap confidence intervals are shown.}
   \label{table:all-methods-cw10-full}
   \centering
 \begin{tabular}{llll}
 \toprule
  method       & performance       & forgetting          & f. transfer   \\
 \midrule
  \textbf{Fine-tuning}            & 0.10 {\tiny [0.10, 0.10]} & 0.75 {\tiny [0.73, 0.76]}   & 0.31 {\tiny [0.27, 0.34]}     \\
  \textbf{Fine-tuning, best-return exploration}        & 0.10 {\tiny [0.10, 0.11]} & 0.73 {\tiny [0.71, 0.75]} & 0.30 {\tiny [0.25, 0.34]}     \\
  \textbf{A-GEM}            & 0.13 {\tiny [0.12, 0.14]} & 0.68 {\tiny [0.66, 0.70]}   & 0.26 {\tiny [0.22, 0.29]}     \\
  \textbf{ClonEx-SAC}          & 0.86 {\tiny [0.84, 0.87]} & 0.02 {\tiny [0.01, 0.04]}   & 0.44 {\tiny [0.42, 0.46]}     \\
  \textbf{Behavioral cloning} & 0.84 {\tiny [0.81, 0.86]} & 0.02 {\tiny [0.01, 0.03]}   & 0.41 {\tiny [0.38, 0.43]}     \\
  \textbf{EWC}             & 0.64 {\tiny [0.60, 0.68]} & 0.06 {\tiny [0.03, 0.09]}   & 0.04 {\tiny [-0.04, 0.12]}    \\
  \textbf{L2}              & 0.53 {\tiny [0.49, 0.58]} & 0.02 {\tiny [-0.00, 0.04]}  & -0.34 {\tiny [-0.47, -0.21]}  \\
  \textbf{MAS}             & 0.53 {\tiny [0.50, 0.57]} & 0.11 {\tiny [0.09, 0.13]}   & -0.06 {\tiny [-0.14, -0.00]}  \\
  \textbf{PackNet}         & 0.84 {\tiny [0.81, 0.86]} & -0.01 {\tiny [-0.02, 0.00]} & 0.26 {\tiny [0.22, 0.29]}     \\
  \textbf{Perfect memory}       & 0.27 {\tiny [0.24, 0.30]} & 0.03 {\tiny [0.00, 0.05]}   & -1.13 {\tiny [-1.23, -1.04]}  \\
  \textbf{VCL}             & 0.55 {\tiny [0.51, 0.59]} & -0.03 {\tiny [-0.05, 0.01]} & -0.37 {\tiny [-0.47, -0.28]}  \\
 \bottomrule
 \end{tabular}
 \end{table}

 \begin{table}[h!]
   \caption{Results of all the methods on CW20 sequence. Average performance, forgetting, and forward transfer are shown in columns. 90\% bootstrap confidence intervals are shown. }
   \label{table:all-methods-cw20-full}
   \centering
 \begin{tabular}{llll}
 \toprule
  method       & performance       & forgetting          & f. transfer   \\
 \midrule
  \textbf{Fine-tuning}            & 0.05 {\tiny [0.05, 0.05]} & 0.74 {\tiny [0.72, 0.76]}   & 0.19 {\tiny [0.15, 0.23]}     \\
  \textbf{Fine-tuning, best-return exploration}        & 0.05 {\tiny [0.05, 0.06]} & 0.74 {\tiny [0.72, 0.76]} & 0.21 {\tiny [0.18, 0.24]}     \\
  \textbf{A-GEM}            & 0.07 {\tiny [0.06, 0.08]} & 0.69 {\tiny [0.68, 0.71]}   & 0.13 {\tiny [0.10, 0.17]}     \\
  \textbf{ClonEx-SAC}          & 0.87 {\tiny [0.86, 0.88]} & 0.02 {\tiny [0.01, 0.03]}   & 0.54 {\tiny [0.52, 0.55]}     \\
  \textbf{Behavioral cloning} & 0.83 {\tiny [0.81, 0.85]} & 0.02 {\tiny [0.01, 0.03]}   & 0.36 {\tiny [0.34, 0.38]}     \\
  \textbf{EWC}             & 0.61 {\tiny [0.59, 0.63]} & 0.03 {\tiny [0.01, 0.05]}   & -0.15 {\tiny [-0.21, -0.09]}  \\
  \textbf{L2}              & 0.50 {\tiny [0.46, 0.53]} & 0.00 {\tiny [-0.01, 0.01]}  & -0.48 {\tiny [-0.59, -0.37]}  \\
  \textbf{MAS}             & 0.53 {\tiny [0.49, 0.57]} & 0.07 {\tiny [0.05, 0.09]}   & -0.16 {\tiny [-0.22, -0.10]}  \\
  \textbf{PackNet}         & 0.80 {\tiny [0.79, 0.82]} & 0.00 {\tiny [-0.01, 0.01]} & 0.18 {\tiny [0.14, 0.22]}     \\
  \textbf{Perfect memory}       & 0.09 {\tiny [0.06, 0.12]} & 0.10 {\tiny [0.08, 0.12]}   & -1.32 {\tiny [-1.41, -1.24]}  \\
  \textbf{VCL}             & 0.50 {\tiny [0.48, 0.53]} & -0.01 {\tiny [-0.03, 0.01]} & -0.45 {\tiny [-0.53, -0.37]}  \\
 \bottomrule
 \end{tabular}
 \end{table}

\section{Additional experiments on transfer in isolation}
\label{sec:experiments_transfer_appendix}

\subsection{Additional matrices}

\begin{figure}[h!]
    \begin{subfigure}[b]{0.325\textwidth}
        \centering
        \includegraphics[width=\textwidth]{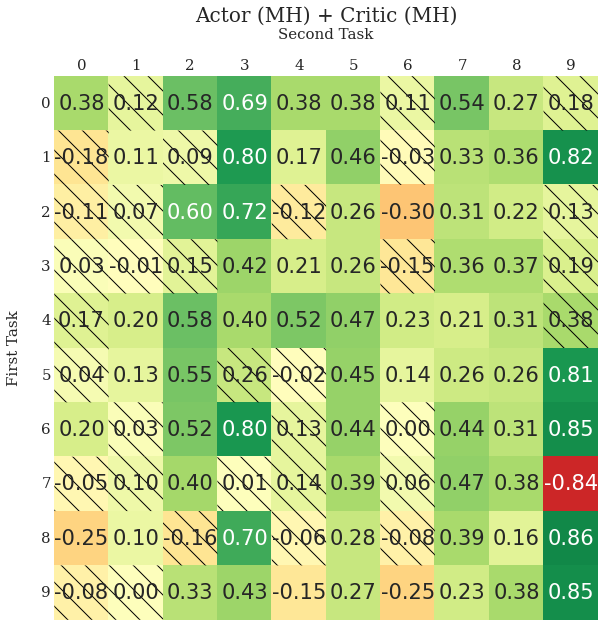}
        \caption{Actor (MH) and Critic (MH) carried over.}
        \label{fig:transfer_actor_mh_critic_mh}
    \end{subfigure}
    \hfill
    \begin{subfigure}[b]{0.325\textwidth}
        \centering
        \includegraphics[width=\textwidth]{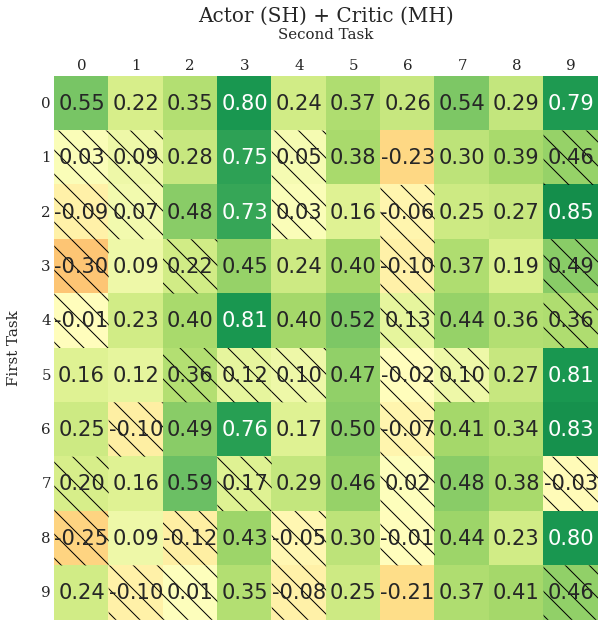}
        \caption{Actor (SH) and Critic (MH) carried over.}
        \label{fig:transfer_actor_sh_critic_mh}
    \end{subfigure}
    \hfill
    \begin{subfigure}[b]{0.325\textwidth}
        \centering
        \includegraphics[width=\textwidth]{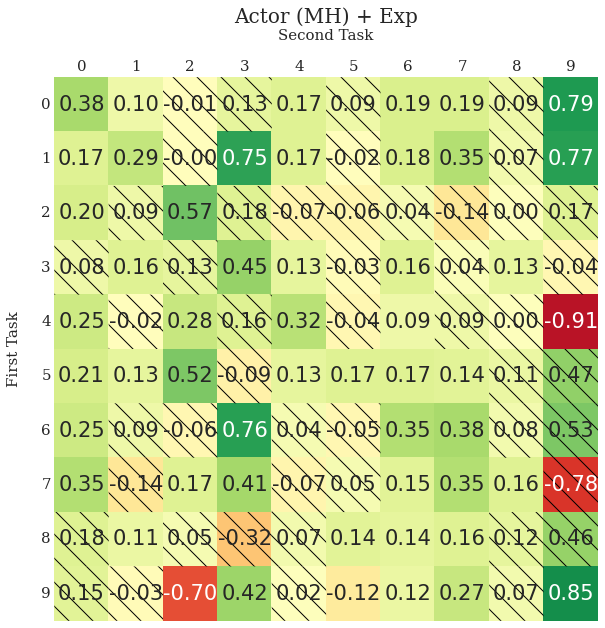}
        \caption{Actor (MH) and exploration carried over.}
        \label{fig:transfer_actor_mh_exp}
    \end{subfigure}

    \hspace*{\fill}%
    \begin{subfigure}[b]{0.325\textwidth}
        \centering
        \includegraphics[width=\textwidth]{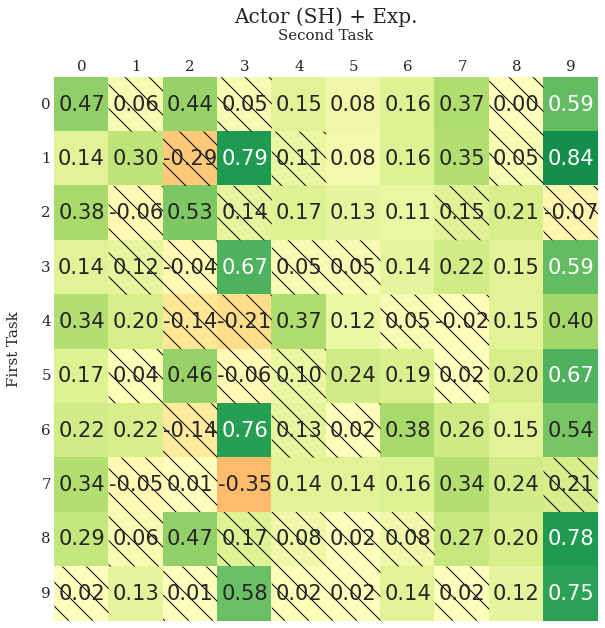}
        \caption{Actor (SH) and exploration carried over.}
        \label{fig:transfer_actor_sh_exp}
    \end{subfigure}
    \hfill
    \begin{subfigure}[b]{0.325\textwidth}
        \centering
        \includegraphics[width=\textwidth]{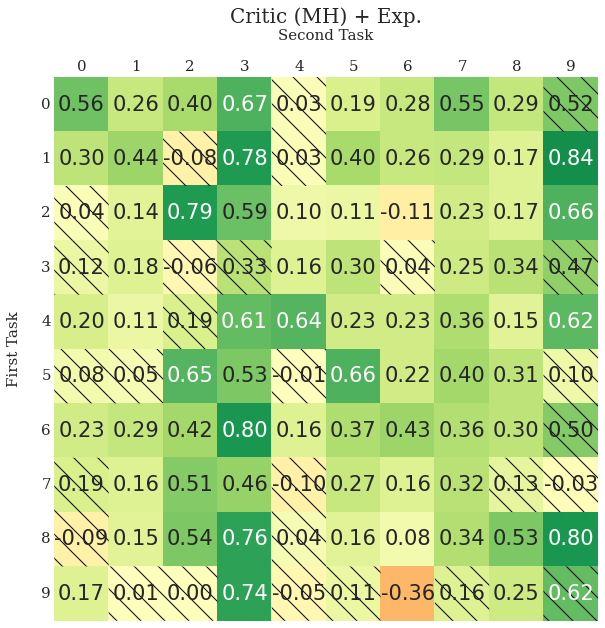}
        \caption{Critic (MH) and exploration carried over}
        \label{fig:transfer_critic_mh_exp}
    \end{subfigure}
    \hspace*{\fill}

    \hspace*{\fill}%
    \begin{subfigure}[b]{0.325\textwidth}
        \centering
        \includegraphics[width=\textwidth]{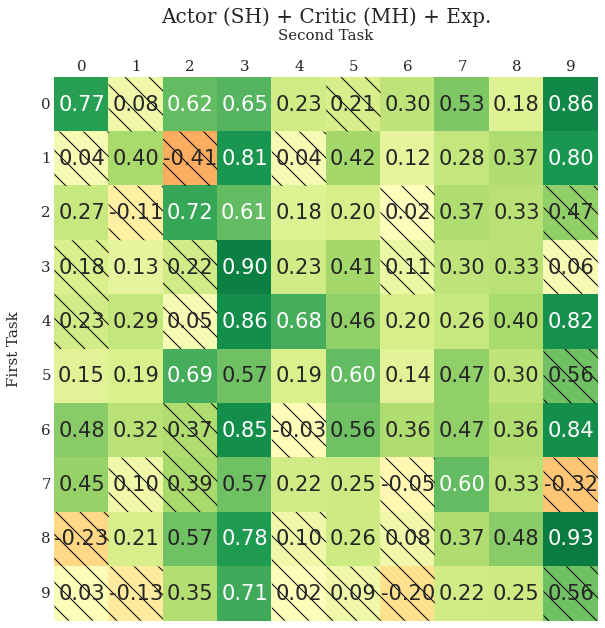}
        \caption{Actor (SH), critic (MH), and exploration carried over.}
        \label{fig:transfer_actor_sh_critic_mh_exp}
    \end{subfigure}
    \hfill
    \begin{subfigure}[b]{0.325\textwidth}
        \centering
        \includegraphics[width=\textwidth]{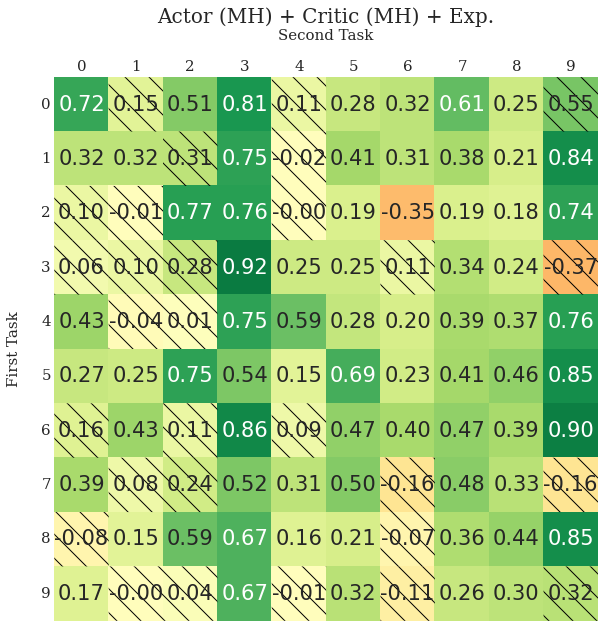}
        \caption{Actor (MH), critic (MH), and exploration carried over.}
        \label{fig:transfer_actor_mh_critic_mh_exp}
    \end{subfigure}
    \hspace*{\fill}
\caption{\small The effect of carrying over different combinations of components on the performance on pairs of tasks from CW10. We shade an entry if the $90\%$ confidence interval contains $0$, indicating that we cannot be sure whether the component which was carried over makes a difference. }
\label{fig:transfer_matrices_combinations}
\end{figure}

In the main text (see Section \ref{sec:transfer_experiments}) we summarized the performance of different combinations of carried over components by calculating the statistics over the whole transfer matrix and presented them in Table \ref{tab:transfer_multiple}. In this section, we provide the full matrices to give a broader picture of the experiments. Figure \ref{fig:transfer_matrices_combinations} contains the result of transferring combinations of components as described previously in the main text. 
As mentioned earlier in Section \ref{sec:basic_transfer}, transferring the actor, the critic, and the exploration policy simultaneously leads to the best results.

\subsection{Transferring the optimizer}

Since we are using the Adam optimizer (Appendix \ref{sec:setting_appendix}), we can also try to transfer the running statistics of the gradients used by the optimizer. However, as we show in Table \ref{tab:appendix_optimizer}, this effect is negligible.

\begin{table}[h!]
    \centering
\begin{tabular}{lccrrr}
\toprule
Setting &                        FT &              FT (no diag) &  \# pos. &  \# neg. &  \# neutral \\
\midrule
Transfer optimizer & 0.02 \tiny{[-0.02, 0.06]} & 0.02 \tiny{[-0.02, 0.06]} &       19 &        5 &          76 \\
\bottomrule
\end{tabular}

\caption{Results when carrying over the optimizer. Its impact 
is negligible.}
\label{tab:appendix_optimizer}
\end{table}

\subsection{Freezing the critic's parameters}

The \textit{Critic (train only head)} row in Table \ref{tab:transfer_basics} shows that simply reusing the features obtained by the penultimate layer of the critic is not enough to achieve good transfer. In fact, the results suggest that in this setting the model is barely able to learn anything at all. In order to investigate this issue, we conduct a series of experiments where we freeze different parts of the critic's backbone. As it consists of $4$ layers (which we number $L1, L2, L3, L4$), we test multiple possible freezing combinations and present the results in Table \ref{tab:appendix_transfer_freezing}. Note that in all experiments the output head is not frozen and can be freely adapted. 
Surprisingly, the results show that freezing the early layers degrades the performance more than freezing the further layers, contradicting the feature reuse theory of transfer learning. We hypothesize that the model instead learns more abstract skills which require rewiring in the early layers rather than the later layers.

\begin{table}[h!]
\centering
\caption{The impact of freezing.}
\begin{tabular}{lccrrr}
\toprule
                 name &                          FT &                FT (no diag) &  \# pos. &  \# neg. &  \# neutral \\
\midrule
Freeze $L1, L2$ & -0.36 \tiny{[-0.41, -0.31]} & -0.42 \tiny{[-0.47, -0.37]} &       18 &       63 &          19 \\
Freeze $L1, L2, L3$ & -0.82 \tiny{[-0.86, -0.78]} & -0.87 \tiny{[-0.91, -0.82]} &        0 &       92 &           8 \\
Freeze $L1, L2, L3, L4$ & -1.29 \tiny{[-1.32, -1.25]} & -1.30 \tiny{[-1.34, -1.27]} &        0 &      100 &           0 \\
Freeze $L2, L3, L4$ & -0.28 \tiny{[-0.32, -0.24]} & -0.29 \tiny{[-0.34, -0.25]} &       14 &       65 &          21 \\
Freeze $L3, L4$ &    0.07 \tiny{[0.03, 0.10]} &    0.06 \tiny{[0.02, 0.10]} &       33 &       27 &          40 \\
\bottomrule
\end{tabular}
\label{tab:appendix_transfer_freezing}
\end{table}

\subsection{Exploration}

By default, the exploration phase of SAC in our experiments takes $10$k steps (see Section \ref{sec:experimental_setup}). In this subsection, we present the results verifying how the length of the exploration phase impacts the outcome. In Table \ref{tab:appendix_exploration_only} we show the results for different exploration phase lengths when only the exploration policy is carried over, and Table \ref{tab:appendix_exploration_actor} shows corresponding results when we carry over both the exploration phase and the actor. As shown, shorter exploration phases usually perform better.

\begin{table}[h!]
\centering
\caption{Effect of the exploration length when carrying over only the exploration policy.}
\begin{tabular}{lccrrr}
\toprule
 Exp. Len. &                         FT &                FT (no diag) &  \# pos. &  \# neg. &  \# neutral \\
\midrule
  500 &   0.06 \tiny{[0.02, 0.09]} &   0.03 \tiny{[-0.01, 0.07]} &       20 &        5 &          75 \\
 1000 &   0.06 \tiny{[0.02, 0.09]} &   0.03 \tiny{[-0.01, 0.07]} &       24 &        3 &          73 \\
 5000 &   0.13 \tiny{[0.10, 0.17]} &    0.10 \tiny{[0.07, 0.14]} &       36 &        5 &          59 \\
10000 &   0.10 \tiny{[0.06, 0.13]} &    0.07 \tiny{[0.03, 0.10]} &       27 &        7 &          66 \\
20000 &   0.10 \tiny{[0.07, 0.14]} &    0.08 \tiny{[0.04, 0.11]} &       30 &       17 &          53 \\
50000 & -0.00 \tiny{[-0.04, 0.03]} & -0.04 \tiny{[-0.08, -0.01]} &       22 &       44 &          34 \\
\bottomrule
\end{tabular}

\label{tab:appendix_exploration_only}
\end{table}

\begin{table}[h!]
\centering
\caption{Effect of exploration length when carrying over the actor's parameters and the exploration policy.}
\begin{tabular}{lccrrr}
\toprule
 Exp. len. &                       FT &             FT (no diag) &  \# pos. &  \# neg. &  \# neutral \\
\midrule
  500 & 0.20 \tiny{[0.16, 0.23]} & 0.17 \tiny{[0.13, 0.21]} &       54 &        1 &          45 \\
 1000 & 0.18 \tiny{[0.15, 0.22]} & 0.16 \tiny{[0.13, 0.20]} &       51 &        0 &          49 \\
 5000 & 0.22 \tiny{[0.19, 0.25]} & 0.20 \tiny{[0.17, 0.24]} &       57 &        1 &          42 \\
10000 & 0.16 \tiny{[0.13, 0.20]} & 0.14 \tiny{[0.10, 0.17]} &       49 &        3 &          48 \\
20000 & 0.15 \tiny{[0.12, 0.19]} & 0.14 \tiny{[0.10, 0.17]} &       43 &        3 &          54 \\
50000 & 0.13 \tiny{[0.10, 0.16]} & 0.11 \tiny{[0.08, 0.15]} &       32 &       12 &          56 \\
\bottomrule
\end{tabular}

\label{tab:appendix_exploration_actor}
\end{table}

\subsection{Transfer independence}
We hint in Section \ref{sec:basic_transfer} the effects of carrying over the actor, critic and exploration add up with regard to the induced transfer. In this section we study this further  by performing a linear regression analysis. Consider $s_{a} \in \{0,1\}$ denoting whether the actor's parameters are carried over ($s_{a} =1$) or not ($s_{a} =0$); analogously, we define $s_c, s_e$ for the critic's parameters and the exploration. The transfer $t(s_a, s_c, s_e)$ reported in Table~\ref{tab:transfer_multiple} fulfills:
$$t(s_a, s_c, s_e) = s_a \cdot w_a + s_c \cdot w_c + s_e \cdot w_e + r(s_a, s_c, s_e),$$
where $w_a = 0.062, w_c = 0.202, w_e = 0.093$, with the residual error $r$ being small; namely $|r(s_a, s_c, s_e)/t(s_a, s_c, s_e)| \leq 0.05$. 

We consider this effect somewhat surprising: there is no synergy nor interference for transferring different components. An explanation of this phenomenon requires further studies. It has an interesting practical implication that one can hope to be able to improve the transfer of any component and assemble it into a 'full' solution.

\section{Additional experiments on transfer in  continual learning.}
\label{sec:experiments_general_cl_appendix}
\subsection{Exploration}
Results for different exploration strategies (see Section~\ref{sec:exploration}) on top of 5 CL methods: Fine-tuning, Behavioral cloning, EWC, L2, PackNet, are presented in Table~\ref{table:exploration-cw10-full} (CW10) and Table~\ref{table:exploration-cw20-full} (CW20). 
We can see that informed exploration strategies help for all considered methods except for PackNet, especially in the forward transfer metric. The effect is more pronounced for the CW20 sequence.

\begin{table}[h!]
  \caption{Results of different exploration strategies added on top of 5 different CL methods, for CW10 sequence. For the description of exploration strategies, see Section~\ref{sec:exploration}.}
  \label{table:exploration-cw10-full}
  \centering
\begin{tabular}{llll}
\toprule
 Method, exploration   & performance       & forgetting        & f. transfer   \\
 \midrule
 \textbf{Fine-tuning, random}               & 0.10 {\tiny [0.10, 0.10]} & 0.74 {\tiny [0.73, 0.76]} & 0.29 {\tiny [0.25, 0.33]}     \\
 \textbf{Fine-tuning, best-return}        & 0.10 {\tiny [0.10, 0.11]} & 0.73 {\tiny [0.71, 0.75]} & 0.30 {\tiny [0.25, 0.34]}     \\
 \textbf{Fine-tuning, preceding}      & 0.10 {\tiny [0.10, 0.11]} & 0.75 {\tiny [0.72, 0.77]} & 0.32 {\tiny [0.27, 0.36]}     \\
 \textbf{Fine-tuning, uniform-previous}   & 0.10 {\tiny [0.10, 0.10]} & 0.75 {\tiny [0.74, 0.77]} & 0.35 {\tiny [0.31, 0.38]}     \\
\midrule
 \textbf{Behavioral cloning, random}               & 0.84 {\tiny [0.81, 0.86]} & 0.02 {\tiny [0.01, 0.03]} & 0.41 {\tiny [0.38, 0.43]}     \\
 \textbf{Behavioral cloning, best-return}        & 0.86 {\tiny [0.84, 0.87]} & 0.02 {\tiny [0.01, 0.04]} & 0.44 {\tiny [0.42, 0.46]}     \\
 \textbf{Behavioral cloning, preceding}      & 0.84 {\tiny [0.82, 0.86]} & 0.02 {\tiny [0.01, 0.03]} & 0.39 {\tiny [0.36, 0.41]}     \\
 \textbf{Behavioral cloning, uniform-previous}   & 0.86 {\tiny [0.85, 0.88]} & 0.01 {\tiny [0.00, 0.03]} & 0.45 {\tiny [0.41, 0.48]}     \\
\midrule
 \textbf{EWC, random}               & 0.63 {\tiny [0.60, 0.66]} & 0.05 {\tiny [0.03, 0.08]}  & 0.03 {\tiny [-0.04, 0.09]}    \\
 \textbf{EWC, best-return}        & 0.70 {\tiny [0.68, 0.73]} & 0.04 {\tiny [0.01, 0.06]}  & 0.25 {\tiny [0.21, 0.28]}     \\
 \textbf{EWC, preceding}      & 0.70 {\tiny [0.67, 0.73]} & 0.02 {\tiny [-0.00, 0.04]} & 0.09 {\tiny [0.03, 0.15]}     \\
 \textbf{EWC, uniform-previous}   & 0.72 {\tiny [0.69, 0.75]} & 0.04 {\tiny [0.02, 0.06]}  & 0.24 {\tiny [0.19, 0.28]}     \\
\midrule
 \textbf{L2, random}               & 0.52 {\tiny [0.48, 0.56]} & 0.01 {\tiny [-0.01, 0.03]}  & -0.47 {\tiny [-0.61, -0.33]}  \\
 \textbf{L2, best-return}        & 0.63 {\tiny [0.60, 0.65]} & -0.01 {\tiny [-0.02, 0.00]} & -0.13 {\tiny [-0.21, -0.05]}  \\
 \textbf{L2, preceding}      & 0.59 {\tiny [0.55, 0.62]} & 0.01 {\tiny [-0.02, 0.05]}  & -0.25 {\tiny [-0.34, -0.16]}  \\
 \textbf{L2, uniform-previous}   & 0.63 {\tiny [0.59, 0.67]} & -0.01 {\tiny [-0.02, 0.01]} & -0.16 {\tiny [-0.28, -0.06]}  \\
\midrule
 \textbf{PackNet, random}               & 0.84 {\tiny [0.81, 0.86]} & -0.01 {\tiny [-0.02, 0.00]}  & 0.26 {\tiny [0.22, 0.29]}     \\
 \textbf{PackNet, best-return}        & 0.85 {\tiny [0.83, 0.86]} & -0.01 {\tiny [-0.02, 0.00]}  & 0.23 {\tiny [0.20, 0.26]}     \\
 \textbf{PackNet, preceding}      & 0.84 {\tiny [0.82, 0.85]} & -0.01 {\tiny [-0.03, -0.00]} & 0.24 {\tiny [0.20, 0.27]}     \\
 \textbf{PackNet, uniform-previous}   & 0.84 {\tiny [0.81, 0.86]} & -0.00 {\tiny [-0.02, 0.01]}  & 0.21 {\tiny [0.15, 0.26]}     \\
\bottomrule
\end{tabular}
\end{table}

\begin{table}[h!]
  \caption{Results of different exploration strategies added on top of 5 different CL methods, for CW20 sequence. For the description of exploration strategies, see Section~\ref{sec:exploration}.}
  \label{table:exploration-cw20-full}
  \centering
\begin{tabular}{llll}
\toprule
 Method, exploration   & performance       & forgetting        & f. transfer   \\
 \midrule
 \textbf{Fine-tuning, random}               & 0.05 {\tiny [0.05, 0.05]} & 0.74 {\tiny [0.73, 0.76]} & 0.20 {\tiny [0.16, 0.23]}     \\
 \textbf{Fine-tuning, best-return}        & 0.05 {\tiny [0.05, 0.06]} & 0.74 {\tiny [0.72, 0.76]} & 0.21 {\tiny [0.18, 0.24]}     \\
 \textbf{Fine-tuning, preceding}      & 0.05 {\tiny [0.05, 0.05]} & 0.76 {\tiny [0.74, 0.78]} & 0.24 {\tiny [0.21, 0.28]}     \\
 \textbf{Fine-tuning, uniform-previous}   & 0.06 {\tiny [0.05, 0.06]} & 0.75 {\tiny [0.74, 0.77]} & 0.24 {\tiny [0.20, 0.27]}     \\
\midrule
 \textbf{Behavioral cloning, random}               & 0.83 {\tiny [0.81, 0.85]} & 0.02 {\tiny [0.01, 0.03]} & 0.36 {\tiny [0.34, 0.38]}     \\
 \textbf{Behavioral cloning, best-return}        & 0.87 {\tiny [0.86, 0.88]} & 0.02 {\tiny [0.01, 0.03]} & 0.54 {\tiny [0.52, 0.55]}     \\
 \textbf{Behavioral cloning, preceding}      & 0.84 {\tiny [0.83, 0.86]} & 0.02 {\tiny [0.01, 0.03]} & 0.40 {\tiny [0.38, 0.42]}     \\
 \textbf{Behavioral cloning, uniform-previous}   & 0.86 {\tiny [0.85, 0.87]} & 0.03 {\tiny [0.02, 0.04]} & 0.51 {\tiny [0.48, 0.53]}     \\
\midrule
 \textbf{EWC, random}               & 0.60 {\tiny [0.59, 0.62]} & 0.03 {\tiny [0.01, 0.04]}  & -0.14 {\tiny [-0.19, -0.09]}  \\
 \textbf{EWC, best-return}        & 0.71 {\tiny [0.69, 0.73]} & 0.01 {\tiny [-0.00, 0.03]} & 0.28 {\tiny [0.25, 0.31]}     \\
 \textbf{EWC, preceding}      & 0.61 {\tiny [0.59, 0.64]} & 0.02 {\tiny [0.00, 0.03]}  & -0.14 {\tiny [-0.19, -0.09]}  \\
 \textbf{EWC, uniform-previous}   & 0.70 {\tiny [0.68, 0.73]} & 0.02 {\tiny [0.01, 0.03]}  & 0.21 {\tiny [0.17, 0.25]}     \\
\midrule
 \textbf{L2, random}               & 0.45 {\tiny [0.41, 0.49]} & 0.02 {\tiny [0.00, 0.05]}   & -0.56 {\tiny [-0.68, -0.45]}  \\
 \textbf{L2, best-return}        & 0.62 {\tiny [0.59, 0.65]} & -0.00 {\tiny [-0.01, 0.01]} & -0.02 {\tiny [-0.09, 0.05]}   \\
 \textbf{L2, preceding}      & 0.48 {\tiny [0.44, 0.52]} & 0.04 {\tiny [0.01, 0.07]}   & -0.47 {\tiny [-0.57, -0.39]}  \\
 \textbf{L2, uniform-previous}   & 0.59 {\tiny [0.56, 0.62]} & 0.00 {\tiny [-0.01, 0.01]}  & -0.15 {\tiny [-0.25, -0.06]}  \\
\midrule
 \textbf{PackNet, random}               & 0.80 {\tiny [0.79, 0.82]} & -0.00 {\tiny [-0.01, 0.01]}  & 0.18 {\tiny [0.14, 0.22]}     \\
 \textbf{PackNet, best-return}        & 0.82 {\tiny [0.81, 0.83]} & -0.00 {\tiny [-0.01, 0.01]}  & 0.23 {\tiny [0.21, 0.25]}     \\
 \textbf{PackNet, preceding}      & 0.81 {\tiny [0.80, 0.83]} & -0.01 {\tiny [-0.02, -0.00]} & 0.20 {\tiny [0.16, 0.24]}     \\
 \textbf{PackNet, uniform-previous}   & 0.80 {\tiny [0.78, 0.82]} & 0.00 {\tiny [-0.01, 0.01]}   & 0.23 {\tiny [0.18, 0.27]}     \\
\bottomrule
\end{tabular}
\end{table}

\subsection{Data rehearsal}

Table~\ref{table:transfer_data_pairs} complements the results of Section~\ref{sec:data_long} and presents the impact of transferring data on the transfer. The data transfer is done either by behavioral cloning or by directly carrying over the SAC buffer. The metrics are computed over 100 pairs of tasks from CW10.
We can see that Behavioral cloning (actor only) does not have an impact on the transfer when training on pairs; the effect becomes visible only for longer sequences (see Table~\ref{table:er-cw10}). Other considered methods harm the transfer.

\begin{table}[t]
\centering

    \caption{Results for different ways of transferring previous data -- behavioral cloning (applied to only the actor, or both the actor and the critic), and directly retaining past tuples in SAC buffer (Transfer RL buffer), on 100 pairs of tasks from CW10 (averaged). Base transfer denotes that the exploration policy and the parameters of the actor and the critic are carried over. FT and FT (no diag) represent average forward transfer across all pairs with and without considering the diagonal (transfer from a task to the same task), respectively. Subsequent columns denote the number of pairs with the positive, negative, and neutral transfer.}
    \label{table:transfer_data_pairs}

\begin{tabular}{lccrrr}
\toprule
name &                          FT &                FT (no diag) &  \# pos. &  \# neg. &  \# neutral \\
\midrule
Base transfer &    0.35 \tiny{[0.31, 0.38]} &    0.32 \tiny{[0.29, 0.36]} &       70 &        1 &          29 \\
\midrule
BC (actor) &   0.01 \tiny{[-0.03, 0.04]} &   0.00 \tiny{[-0.03, 0.04]} &       28 &        9 &          63 \\
BC (actor) + base transfer &    0.37 \tiny{[0.34, 0.40]} &    0.35 \tiny{[0.32, 0.38]} &       77 &        2 &          21 \\
BC (actor+critic) & -0.56 \tiny{[-0.61, -0.52]} & -0.62 \tiny{[-0.67, -0.58]} &        3 &       77 &          20 \\
BC (actor+critic) + base transfer &  -0.04 \tiny{[-0.08, 0.01]} & -0.12 \tiny{[-0.16, -0.07]} &       35 &       34 &          31 \\

\midrule

Transfer RL buffer & -0.82 \tiny{[-0.86, -0.78]} & -0.90 \tiny{[-0.94, -0.85]} &        4 &       88 &           8 \\
Transfer RL buffer + base transfer & -0.59 \tiny{[-0.63, -0.54]} & -0.71 \tiny{[-0.76, -0.66]} &       12 &       75 &          13 \\

\bottomrule
\end{tabular}

\end{table}

\subsection{Regularizing the critic}
\label{app:reg-critic}

We vary the coefficient for critic regularization and measure its impact on final metrics for three different methods: Behavioral cloning, EWC, and L2. We set the actor's regularization weight to the optimal value and run a sweep over the regularization coefficient of the critic. We run experiments on the CW10 sequence. The results are presented in Tables~\ref{table:cw10-bc-critic-reg},~\ref{table:cw10-ewc-critic-reg},~\ref{table:cw10-l2-critic-reg}, indicating that direct regularization of the critic does not significantly improve the performance.

\begin{table}
\centering
\caption{Average performance and forward transfer metrics on CW10 for Behavioral cloning, for different values of the critic regularization coefficient.}
\label{table:cw10-bc-critic-reg}
\begin{tabular}{rll}
\toprule
   critic’s regularization coef. & performance       & f. transfer   \\
\midrule
            0      & 0.82 {\tiny [0.80, 0.83]} & 0.34 {\tiny [0.30, 0.37]}     \\
            1e-10  & 0.83 {\tiny [0.81, 0.85]} & 0.36 {\tiny [0.33, 0.39]}     \\
            1e-09  & 0.83 {\tiny [0.81, 0.84]} & 0.36 {\tiny [0.32, 0.39]}     \\
            1e-08  & 0.79 {\tiny [0.78, 0.81]} & 0.32 {\tiny [0.28, 0.35]}     \\
            1e-07  & 0.83 {\tiny [0.81, 0.84]} & 0.35 {\tiny [0.32, 0.38]}     \\
            1e-06  & 0.81 {\tiny [0.80, 0.83]} & 0.37 {\tiny [0.36, 0.39]}     \\
            1e-05  & 0.81 {\tiny [0.79, 0.82]} & 0.37 {\tiny [0.35, 0.39]}     \\
            0.0001 & 0.82 {\tiny [0.81, 0.84]} & 0.39 {\tiny [0.36, 0.41]}     \\
            0.001  & 0.77 {\tiny [0.75, 0.78]} & 0.33 {\tiny [0.30, 0.35]}     \\
            0.01   & 0.71 {\tiny [0.69, 0.73]} & 0.23 {\tiny [0.20, 0.26]}     \\
            0.1    & 0.68 {\tiny [0.66, 0.69]} & 0.11 {\tiny [0.07, 0.14]}     \\
            1      & 0.65 {\tiny [0.63, 0.66]} & -0.02 {\tiny [-0.06, 0.02]}   \\
           10      & 0.62 {\tiny [0.60, 0.63]} & -0.18 {\tiny [-0.22, -0.14]}  \\
          100      & 0.48 {\tiny [0.46, 0.50]} & -0.58 {\tiny [-0.64, -0.53]}  \\
\bottomrule
\end{tabular}
\end{table}

\begin{table}
\centering
\caption{Average performance and forward transfer metrics on CW10 for EWC, for different values of the critic regularization coefficient.}
\label{table:cw10-ewc-critic-reg}
\begin{tabular}{rll}
\toprule
   critic’s regularization coef. & performance       & f. transfer   \\
\midrule
            0      & 0.66 {\tiny [0.64, 0.67]} & 0.08 {\tiny [0.04, 0.11]}     \\
            1e-10  & 0.64 {\tiny [0.62, 0.66]} & 0.05 {\tiny [0.01, 0.09]}     \\
            1e-09  & 0.64 {\tiny [0.62, 0.66]} & 0.06 {\tiny [0.02, 0.10]}     \\
            1e-08  & 0.62 {\tiny [0.60, 0.64]} & 0.06 {\tiny [0.02, 0.10]}     \\
            1e-07  & 0.62 {\tiny [0.59, 0.64]} & 0.06 {\tiny [0.01, 0.10]}     \\
            1e-06  & 0.63 {\tiny [0.61, 0.65]} & 0.01 {\tiny [-0.04, 0.06]}    \\
            1e-05  & 0.63 {\tiny [0.60, 0.65]} & 0.02 {\tiny [-0.02, 0.05]}    \\
            0.0001 & 0.61 {\tiny [0.59, 0.63]} & -0.03 {\tiny [-0.07, 0.01]}   \\
            0.001  & 0.55 {\tiny [0.53, 0.57]} & -0.10 {\tiny [-0.15, -0.05]}  \\
            0.01   & 0.50 {\tiny [0.47, 0.52]} & -0.25 {\tiny [-0.31, -0.19]}  \\
            0.1    & 0.40 {\tiny [0.37, 0.42]} & -0.50 {\tiny [-0.56, -0.44]}  \\
            1      & 0.27 {\tiny [0.25, 0.29]} & -0.85 {\tiny [-0.92, -0.79]}  \\
           10      & 0.18 {\tiny [0.16, 0.19]} & -1.24 {\tiny [-1.32, -1.18]}  \\
          100      & 0.12 {\tiny [0.10, 0.13]} & -1.45 {\tiny [-1.52, -1.39]}  \\
        10000      & 0.11 {\tiny [0.10, 0.13]} & -1.45 {\tiny [-1.53, -1.38]}  \\
\bottomrule
\end{tabular}
\end{table}

\begin{table}
\centering
\caption{Average performance and forward transfer metrics on CW10 for L2, for different values of the critic regularization coefficient.}
\label{table:cw10-l2-critic-reg}
\begin{tabular}{rll}
\toprule
   critic’s regularization coef. & performance       & f. transfer   \\
\midrule
            0      & 0.53 {\tiny [0.50, 0.56]} & -0.40 {\tiny [-0.48, -0.33]}  \\
            1e-10  & 0.53 {\tiny [0.51, 0.55]} & -0.37 {\tiny [-0.44, -0.32]}  \\
            1e-09  & 0.55 {\tiny [0.53, 0.58]} & -0.35 {\tiny [-0.42, -0.28]}  \\
            1e-08  & 0.53 {\tiny [0.50, 0.56]} & -0.34 {\tiny [-0.41, -0.28]}  \\
            1e-07  & 0.52 {\tiny [0.50, 0.55]} & -0.38 {\tiny [-0.45, -0.32]}  \\
            1e-06  & 0.54 {\tiny [0.51, 0.56]} & -0.41 {\tiny [-0.49, -0.35]}  \\
            1e-05  & 0.55 {\tiny [0.52, 0.57]} & -0.36 {\tiny [-0.43, -0.29]}  \\
            0.0001 & 0.52 {\tiny [0.49, 0.56]} & -0.47 {\tiny [-0.56, -0.39]}  \\
            0.001  & 0.53 {\tiny [0.50, 0.55]} & -0.44 {\tiny [-0.52, -0.37]}  \\
            0.01   & 0.53 {\tiny [0.50, 0.55]} & -0.41 {\tiny [-0.50, -0.34]}  \\
            0.1    & 0.49 {\tiny [0.46, 0.52]} & -0.45 {\tiny [-0.54, -0.36]}  \\
            1      & 0.49 {\tiny [0.46, 0.52]} & -0.44 {\tiny [-0.53, -0.35]}  \\
           10      & 0.50 {\tiny [0.46, 0.53]} & -0.40 {\tiny [-0.50, -0.31]}  \\
          100      & 0.45 {\tiny [0.43, 0.48]} & -0.49 {\tiny [-0.57, -0.41]}  \\
        10000      & 0.25 {\tiny [0.23, 0.27]} & -1.08 {\tiny [-1.17, -1.01]}  \\
       100000      & 0.13 {\tiny [0.12, 0.15]} & -1.41 {\tiny [-1.48, -1.35]}  \\
\bottomrule
\end{tabular}
\end{table}

\section{Infrastructure}
\label{sec:infra_appendix}

In our experiments, we use CPU servers, provided through a cloud service. Throughout the whole project, we conducted over $100.000$ runs with 12 cores per run and an average of $10$ hours per run, which in the end sums up to over $12$M CPU hours.

\section{Continual World benchmark}

\begin{figure}
    \vskip -1in
    \begin{center}
        \centerline{
        \includegraphics[width=.9\columnwidth]{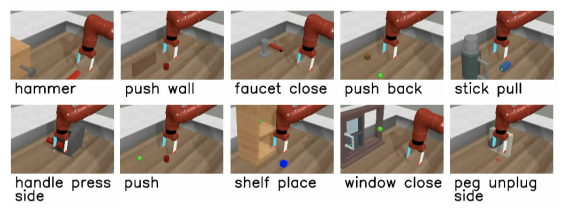}
        }
        \caption{\small Continual World benchmark adopts robotic tasks from Meta-World benchmark. Depicted above is the CW10 sequence. The CW20 sequence contains tasks from CW10 repeated twice. Tasks are trained sequentially, each one for 1M steps.}
        \label{fig:cw20}
    \end{center}
    \vskip -0.3in
  \end{figure}

We briefly present the Continual World benchmark in Figure~\ref{fig:cw20}.

\end{document}